\PassOptionsToPackage{svgnames,table,dvipsnames}{xcolor}
\documentclass{article} % For LaTeX2e
\usepackage{iclr2023_conference,times}
\usepackage[frozencache,cachedir=.]{minted}
\usepackage{rotating}

\usepackage{amsmath,amsfonts,bm}

\def\eqref#1{equation~\ref{#1}}
\def\1{\bm{1}}

\DeclareMathAlphabet{\mathsfit}{\encodingdefault}{\sfdefault}{m}{sl}
\SetMathAlphabet{\mathsfit}{bold}{\encodingdefault}{\sfdefault}{bx}{n}

\usepackage{multirow}
\usepackage{makecell}

\usepackage{hyperref}
\usepackage{url}
\usepackage{wrapfig}
\usepackage{graphicx}
\usepackage[export]{adjustbox}% for the second solution
\usepackage{subfigure}
\usepackage{algorithm}
\usepackage{algpseudocode}
\usepackage{bbold}
\usepackage{xcolor}

\usepackage{tabularx}
\usepackage{colortbl}
\usepackage{booktabs}
\usepackage{cellspace}

\definecolor{LightCyan}{rgb}{0.93,0.947,0.985}
\definecolor{LightGold}{rgb}{1,0.93,0.85}
\definecolor{LightGrey}{rgb}{0.93,0.93,0.93}
\definecolor{darkblue}{rgb}{0.3,0.3,0.6}

\newcolumntype{a}{>{\columncolor{LightCyan}}c}
\newcolumntype{b}{>{\columncolor{LightGold}}c}
\newcolumntype{g}{>{\columncolor{LightGrey}}c}

\usepackage{ifthen}
\usepackage{amssymb}
\newboolean{showcomments}
\ifthenelse{\boolean{showcomments}}
  {\newcommand{\nb}[2]{
    \fcolorbox{gray}{yellow}{\bfseries\sffamily\scriptsize#1}
    {\sf\small$\blacktriangleright$\textit{#2}$\blacktriangleleft$}
   }
   
  }
  {\newcommand{\nb}[2]{}
   
  }

\newcommand{\todonr}[1]{\smash{\fcolorbox{gray}{red}{?}}}

\title{Understanding the Covariance Structure \\ of Convolutional Filters}

\author{%
Asher Trockman$^1$, Devin Willmott$^2$, J. Zico Kolter$^{12}$  \\
$^1$Carnegie Mellon University and $^2$Bosch Center for AI \\
Correspondence to: \texttt{ashert@cs.cmu.edu}
}

\iclrfinalcopy % Uncomment for camera-ready version, but NOT for submission.
\begin{document}

\maketitle

\looseness=-1
\begin{abstract}
Neural network weights are typically initialized at random from univariate distributions,
controlling just the variance of individual weights even in highly-structured operations like convolutions. Recent ViT-inspired convolutional networks such as ConvMixer and ConvNeXt use large-kernel depthwise convolutions whose learned filters have notable structure; this presents an opportunity to study their empirical covariances. In this work, we first observe that such learned filters have highly-structured covariance matrices, and moreover, we find that covariances calculated from small networks may be used to effectively initialize a variety of larger networks of different depths, widths, patch sizes, and kernel sizes, indicating a degree of model-independence to the covariance structure. Motivated by these findings, we then propose a learning-free \emph{multivariate} initialization scheme for convolutional filters using a simple, closed-form construction of their covariance. Models using our initialization outperform those using traditional univariate initializations, and typically meet or exceed the performance of those initialized from the covariances of learned filters; in some cases, this improvement can be achieved \emph{without training the depthwise convolutional filters at all}.
\end{abstract}

\section{Introduction}

Early work in deep learning for vision demonstrated that the convolutional filters
in trained neural networks are often highly-structured, in some cases being
qualitatively similar to filters known from classical computer vision~\citep{alex}.
However, for many years it became standard to replace large-filter convolutions with 
stacked small-filter convolutions, which have less room for any notable
amount of structure.
But in the past year, this trend has changed with inspiration from
the long-range spatial mixing abilities of vision transformers. 
Some of the most  prominent new convolutional neural networks, such as ConvNeXt and ConvMixer,
once again use large-filter convolutions.
These new models also completely separate the processing
of the channel and spatial dimensions, meaning that the now-single-channel filters are,
in some sense, more independent from each other than in previous models such as ResNets.
This presents an opportunity to investigate the structure of convolutional filters.

In particular, we seek to understand the \emph{statistical structure} of convolutional filters,
with the goal of more effectively initializing them.
Most initialization strategies for neural networks focus simply on controlling the \emph{variance}
of weights, as in Kaiming~\citep{kaiming} and Xavier~\citep{xavier} initialization,
which neglect the fact that many layers in neural networks are highly-structured,
with interdependencies between weights, particularly after training.
Consequently, we study the \emph{covariance} matrices of the parameters of convolutional filters,
which we find to have a large degree of perhaps-interpretable structure.
We observe that the covariance of filters calculated from pre-trained models 
can be used to effectively initialize new convolutions by sampling filters from the corresponding
multivariate Gaussian distribution.

We then propose a closed-form and completely learning-free construction of covariance
matrices for randomly initializing convolutional filters from Gaussian distributions.
Our initialization is \emph{highly effective}, especially for larger filters, deeper models, and
shorter training times;
it usually outperforms both standard uniform
initialization techniques \emph{and} our baseline technique of initializing by sampling from the distributions of
pre-trained filters,
both in terms of final accuracy and time-to-convergence.
Models using our initialization often see gains of over 1\% accuracy
on CIFAR-10 and short-training ImageNet classification;
it also leads to small but significant performance gains on full-scale,
$\approx80\%$-accuracy ImageNet training.
Indeed, in some cases our initialization works so well that it outperforms
uniform initialization \emph{even when the filters aren't trained at all}.
And our initialization is almost completely \emph{free to compute.}

\paragraph{Related work}
\cite{saxeortho} proposed to replace random \emph{i.i.d.} Gaussian weights 
with random orthogonal matrices,
a constraint in which weights depend on each other and are thus, in some sense, ``multivariate'';
\cite{initortho} also proposed an orthogonal initialization for convolutions.
Similarly to these works, our initialization
greatly improves the trainability of deep (depthwise) convolutional networks,
but is much simpler and constraint-free, being just
a random sample from a multivariate Gaussian distribution.
\cite{kernelshaping} uses ``Gaussian Delta initialization'' for convolutions;
while largely unrelated to our technique both in form and motivation,
this is similar to our initialization as applied in the first layer
(\emph{i.e.,} the lowest-variance case).
\cite{lego} suggests that the main purpose of pre-training 
may be to find a good initialization,
and crafts a \emph{mimicking initialization}
based on observed, desirable information transfer patterns.
We similarly initialize convolutional filters to be closer to those
found in pre-trained models, but do so in a completely random and simpler manner.
\cite{flexconv} proposes an analytic parameterization of variable-size convolutions,
based in part on Gaussian filters;
while our covariance construction is also analytic and built upon Gaussian filters, we use them to specify
the \emph{distribution} of filters.

Our contribution is most advantageous
for large-filter convolutions, which have become prevalent in recent work:
ConvNeXt~\citep{convnext} uses $7\times7$ convolutions, and ConvMixer~\citep{trockman} uses $9\times 9$;
taking the trend a step further, \cite{ding2022scaling} uses $31\times31$, and
\cite{liu2022more} uses $51\times 51$ sparse convolutions.
Many other works argue for large-filter convolutions~\citep{wang2022can, chen2022scaling, han2021connection}.

\paragraph{Preliminaries}
This work is concerned with depthwise convolutional filters, each of which is parametrized by a $k \times k$ matrix, where $k$ (generally odd) denotes the filter's size. Our aim is to study distributions that arise from convolutional filters in pretrained networks, and to explore properties of distributions whose samples produce strong initial parameters for convolutional layers. More specifically, we hope to understand the covariance among pairs of filter parameters for fixed filter size $k$.
This is intuitively expressed as a covariance matrix $\Sigma \in \mathbb{R}^{k^2 \times k^2}$ with block structure: $\Sigma$ has $k \times k$ blocks, where each block $\left[ \Sigma_{i,j} \right] \in \mathbb{R}^{k \times k}$ corresponds to the covariance between filter pixel $i,j$ and all other $k^2 - 1$ filter pixels.
That is, $\left[ \Sigma_{i,j}\right]_{\ell, m} = \left[ \Sigma_{\ell, m}\right]_{i,j}$ gives the covariance of pixels $i,j$ and $\ell, m$.

In practice, we restrict our study to multivariate Gaussian distributions, which by convention are considered as distributions over $n$-dimensional \textit{vectors} rather than matrices, where the distribution $\mathcal{N}(\mu, \Sigma')$ has a covariance matrix $\Sigma' \in \mathbb{S}^n_+$ where $\Sigma'_{i,j} = \Sigma'_{j,i}$ represents the covariance between vector elements $i$ and $j$.
To align with this convention when sampling filters, we convert from our original block covariance matrix representation to the representation above by simple reassignment of matrix entries, given by
\begin{equation}
\Sigma'_{ki+j, k\ell+m} := \left[ \Sigma_{i,j} \right]_{\ell, m} \text{ for } 1 \leq i, j, \ell, m \leq k
    \label{eq:rearr}
\end{equation}
or, equivalently,
\begin{equation}
\Sigma'_{ki+j, :} := \text{vec}\left(\left[ \Sigma_{i,j} \right]\right) \text{ for } 1 \leq i, j \leq k.
\end{equation}
In this form, we may now generate a filter $F \in \mathbb{R}^{k \times k}$ by drawing a sample $f \in \mathbb{R}^{k^2}$ from $\mathcal{N}(\mu, \Sigma')$ and assigning $F_{i,j} := f_{ki+j}$. Throughout the paper, we assume covariance matrices are in the block form unless we are sampling from a distribution, where the conversion between forms is assumed.

\paragraph{Scope}
We restricted our study to networks
made by stacking simple blocks which each have a single \emph{depthwise} convolutional layer (that is, filters in the layer act on each input channel separately, rather than summing features over input channels), plus other operations such as
pointwise convolutions or MLPs; the \emph{depth} of networks throughout the paper is
synonymous with the number of depthwise convolutional layers,
though this is not the case for neural networks more generally.
All networks investigated use a fixed filter size throughout the network, though the methods
we present could easily be extended to the non-uniform case.
Further, all methods presented do not concern the biases of convolutional layers.

\section{The Covariances of Trained Convolutional Filters and Their Transferability Across Architectures}
\label{sec:cov-transfer}

In this section, we propose a simple starting point in our investigation of convolutional filter
covariance structure:
using the distribution of filters from \emph{pre-trained models} to initialize
filters in new models, a process we term \emph{covariance transfer}.
In the simplest case, we use a pre-trained model with exactly the same architecture as the model to be initialized;
we then show that we can actually transfer filter covariances across very different models.

\paragraph{Basic method.} We use $i\in 1, \dots, D$ to denote the $i^{\text{th}}$ depthwise convolutional layer of a model with $D$ layers.
We denote the $j \in 1, \dots, H$ filters of the $i^{\text{th}}$ pre-trained layer of the model by 
$F_{ij}$ for a model with $H$ convolutional filters in a particular layer (\emph{i.e.,} hidden dimension $H$)
and $F^\prime$ to denote the filters of a new, untrained model.
Then the empirical covariance of the filters in layer $i$ is%
\begin{equation}
    \Sigma_i = \text{Cov}[ \text{vec}(F_{i1}), \ldots, \text{vec}(F_{iH})],
\end{equation}
with the mean $\mu_i$ computed similarly.
Then the new model can be initialized by drawing filters from the multivariate Gaussian distribution
with parameters $\mu_i, \Sigma_i$:
\begin{equation}
    F^\prime_{ij} \sim \mathcal{N}(\mu_i, \Sigma_i) \quad \text{for } j \in 1,\dots,H, i \in 1,\dots,D
\end{equation}
Note that in this section, we use the means of the filters in addition to the covariances
to define the distributions from which to initialize.
However, we found that the mean can be assumed to be zero with little change in performance,
and we focus solely on the covariance in later sections.

\begin{figure}
  \begin{minipage}[c]{0.33\textwidth}
    \caption{In pre-trained models, the covariance
        matrices of convolutional filters
        are \emph{highly-structured.}
        Filters in earlier layers tend to be focused,
        becoming more diffuse as depth increases.
        Observing the structure of each sub-block,
        we note that there is often a static, centered
        negative component and a dynamic positive component
        that moves according to the block's position.
        Often, covariances are higher towards the center of
        the filters.
    } \label{fig:cifar-p1-covs}
  \end{minipage}
  \begin{minipage}[c]{0.67\textwidth}
    \includegraphics[width=\textwidth]{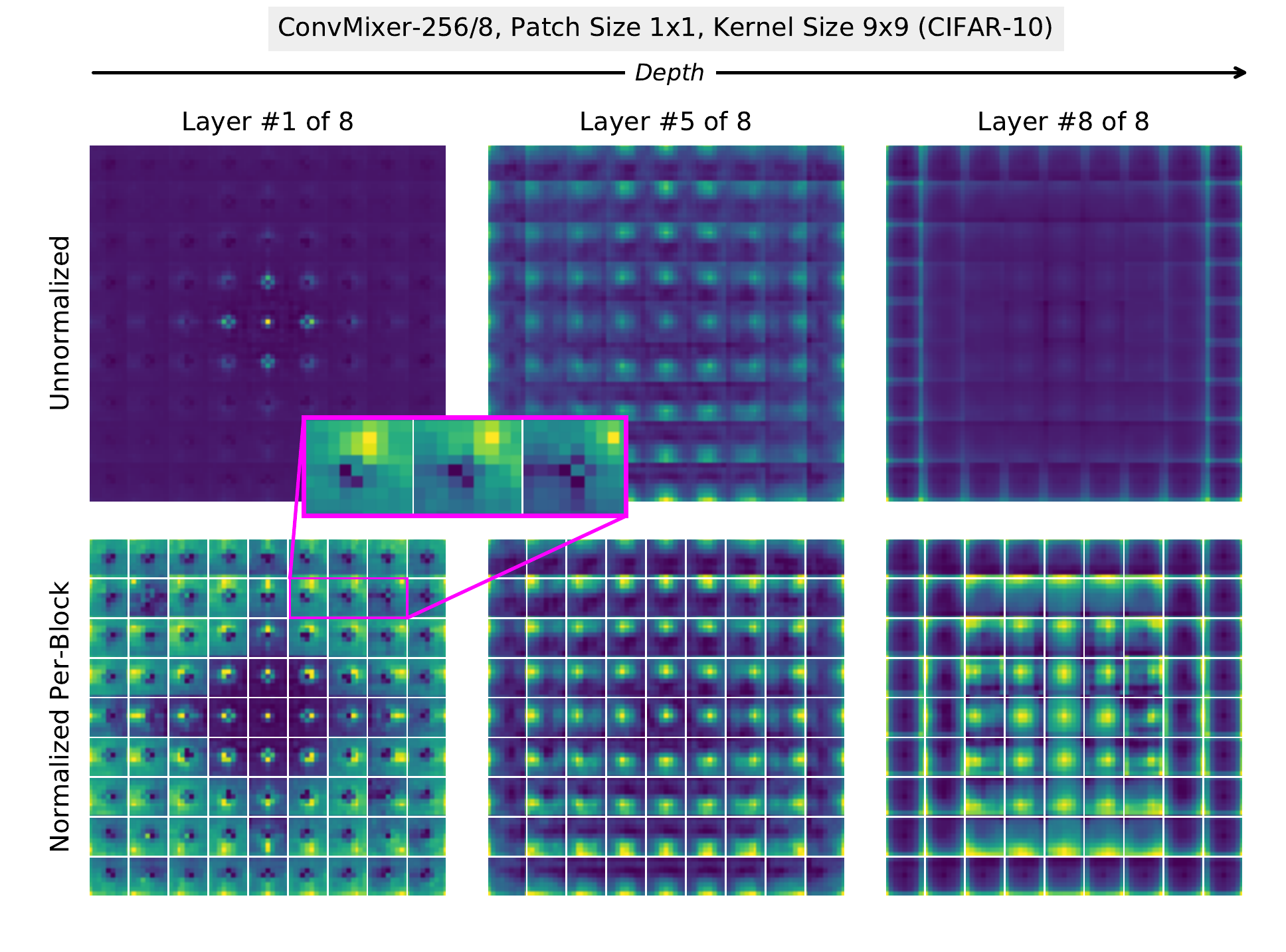}
  \end{minipage}\hfill
\end{figure}

\paragraph{Experiment design.}
We test our initialization methods
primarily on ConvMixer
since it is simple and exceptionally easy to train on CIFAR-10.
We use FFCV~\citep{ffcv} for fast data loading
using our own implementations of fast depthwise convolution
and RandAugment~\citep{randaug}.
To demonstrate the performance of our methods across a variety of
training times, we train for 20, 50, or 200 epochs with a batch size of 512,
and we repeat all experiments with three random seeds.
For all experiments, we use a simple triangular learning rate schedule with the AdamW optimizer,
a learning rate of .01, and weight decay of .01 as in \cite{trockman}.

Most of our CIFAR experiments use a ConvMixer-256/8 with either patch size 1 or 2;
a ConvMixer-$H$/$D$ has precisely $D$ depthwise convolutional layers with $H$ filters each,
ideal for testing our initial covariance transfer techniques.
We train ConvMixers using popular filter sizes 3, 7, and 9, as well as 15
(see Appendix~\ref{apx:cifar} for 5).
We also test our methods on ConvNeXt~\citep{convnext},
which includes downsampling unlike ConvMixer;
we use a patch size of 1 or 2 with ConvNeXt rather than the default 4
to accomodate relatively small CIFAR-10 images, and the default $7 \times 7$ filters.

For most experiments, we provide two baselines for comparison: standard uniform
initialization, the standard in PyTorch~\citep{kaiming},
as well as \emph{directly} transferring the learned filters from a pre-trained model
to the new model.
In most cases, we expect new random initializations to fall between the performance
of uniform and direct transfer initializations.
For our covariance transfer experiments, we trained
a variety of reference models from which to compute covariances; 
these are all trained for the full 200 epochs using the same settings as above.

\paragraph{Frozen filters.}
\cite{rethinking} noticed that ConvMixers with $3\times3$ filters perform well
\begin{wrapfigure}[11]{r}{0.3\textwidth}
  \vspace{-1em}
  \begin{center}
    \includegraphics[width=0.3\textwidth,trim=5 5 5 5,clip]{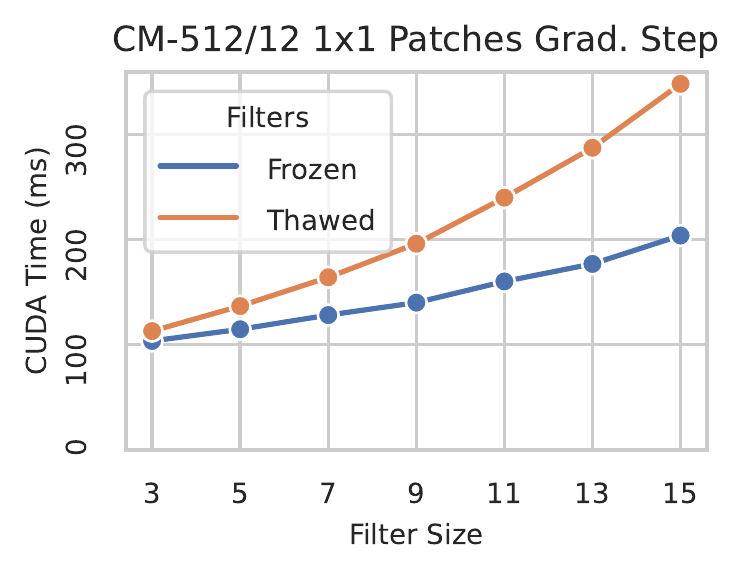}
  \end{center}
	\vspace*{-1em}
	\caption{The backward pass is faster with frozen filters.}
  \label{fig:fast-freeze}
\end{wrapfigure}
even when the filters are \emph{frozen}; that is, the filter weights
remain unchanged over the course of training, receiving no gradient updates.
As we are initializing filters from the distribution of trained filters,
we suspect that additional training may not be completely necessary.
Consequently, in all experiments we investigate both models with \emph{thawed}
filters as well as their \emph{frozen} counterparts.
Freezing filters removes one of the two
gradient calculations from depthwise convolution, resulting in substantial
training speedups as kernel size increases (see Figure~\ref{fig:fast-freeze}).
ConvMixer-512/12 with kernel size $9 \times 9$ is around 20\% faster, while $15 \times 15$ is around 40\% faster.
Further, good performance in the frozen filter setting suggests
that an initialization technique is highly effective.

\subsection{Results}

\begin{figure}[t]
    \centering
    \subfigure{\includegraphics[width=\textwidth,trim=7 20 7 5,clip]{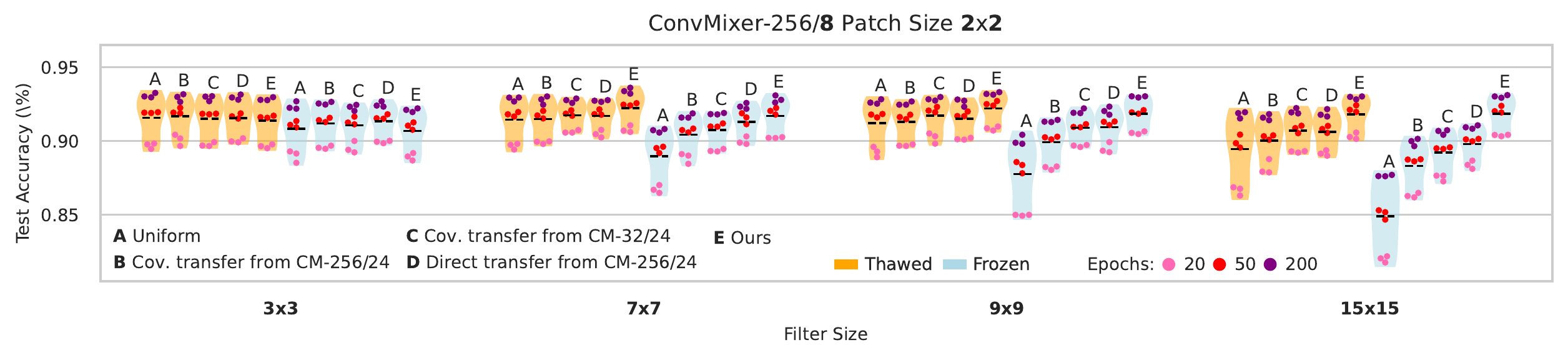}}
    \subfigure{\includegraphics[width=\textwidth,trim=7 20 7 5,clip]{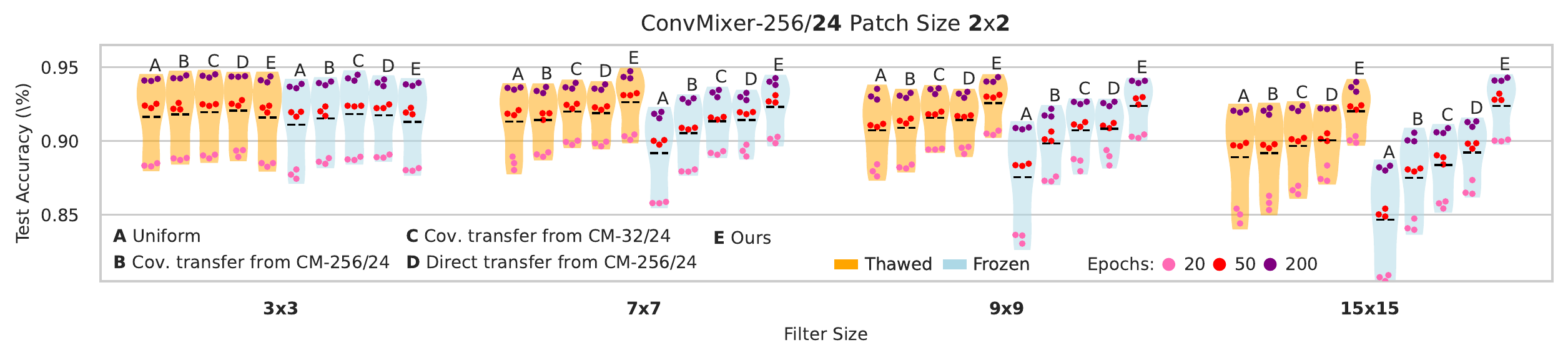}}
    \subfigure{\includegraphics[width=\textwidth,trim=7 5 7 5,clip]{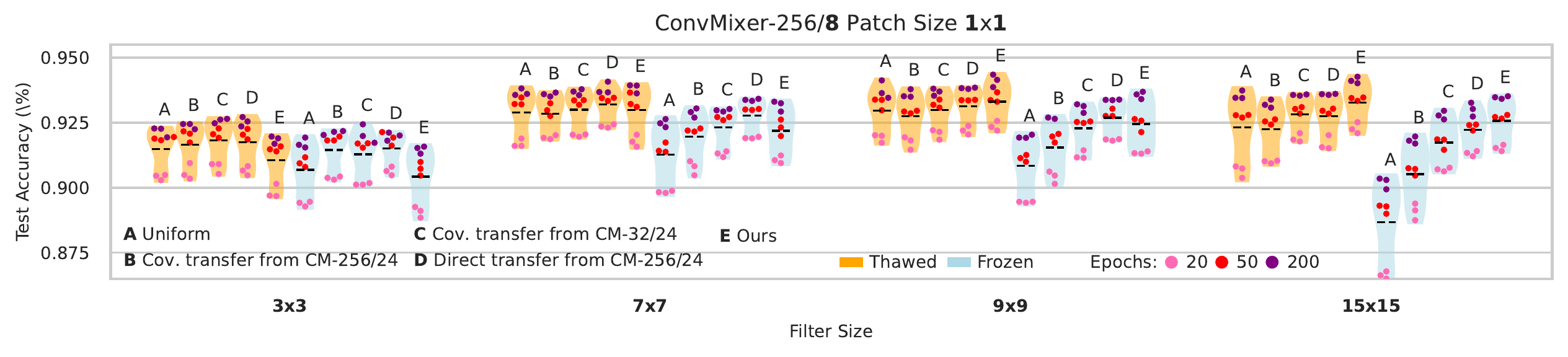}}
    \vspace*{-1.5em}
    \caption{CIFAR-10 accuracy for uniform initialization (\textbf{A}), baseline covariance transfer (\textbf{B-D}), and our custom initialization results (\textbf{E}).}
    \label{fig:cifar-minit}
\end{figure}

The simplest case of covariance transfer (from exactly the same architecture)
is a fairly effective initialization scheme for convolutional filters.
In Fig.~\ref{fig:cifar-minit},
note that this case of covariance transfer (group \textbf{B})
results in somewhat higher accuracies than uniform initialization (group \textbf{A}),
particularly for 20-epoch training;
it also substantially improves the case for frozen filters.
Across all trials, the effect of using this initialization
is higher for larger kernel sizes.
In Fig.~\ref{fig:minit-converge}, we show that covariance
transfer (\emph{gold}) initially increases convergence, but the 
advantange over uniform initialization quickly fades.
As expected, covariance transfer tends to fall between the performance
of \emph{direct transfer}, where we directly initialize using the filters
of the pre-trained model, and default uniform initialization
(see group \textbf{D} in Fig.~\ref{fig:cifar-minit} and the \emph{green}
curves in Fig.~\ref{fig:minit-converge}).

However, we acknowledge that it is not appealing to pre-train models
just for an initialization technique with rather marginal gains,
so we explore the feasibility of covariance transfer from \emph{smaller} models,
both in terms of width and depth.

\begin{figure}[t]
	\centering
    \subfigure{\includegraphics[width=\textwidth,trim=7 20 7 5,clip]{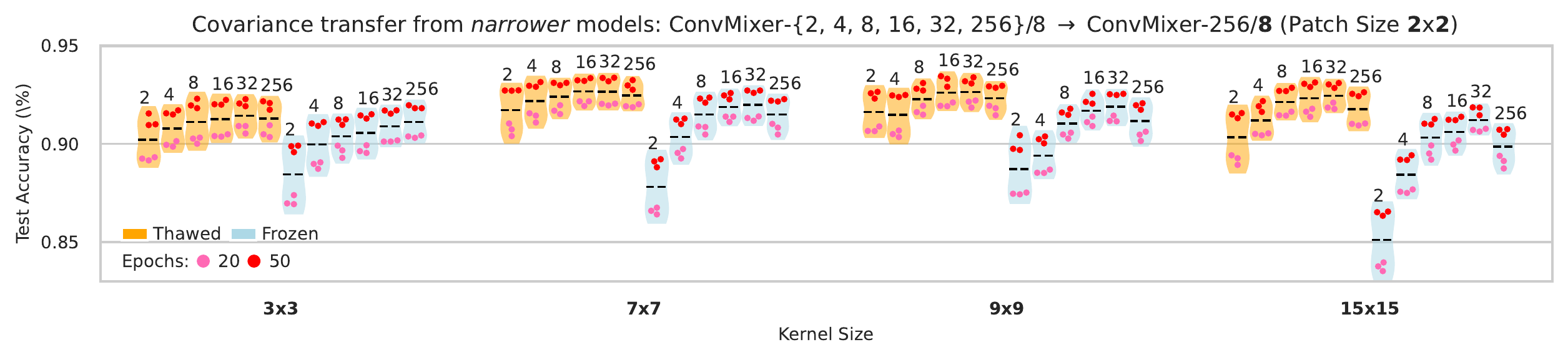}}
	\subfigure{\includegraphics[width=\textwidth,trim=7 5 7 5,clip]{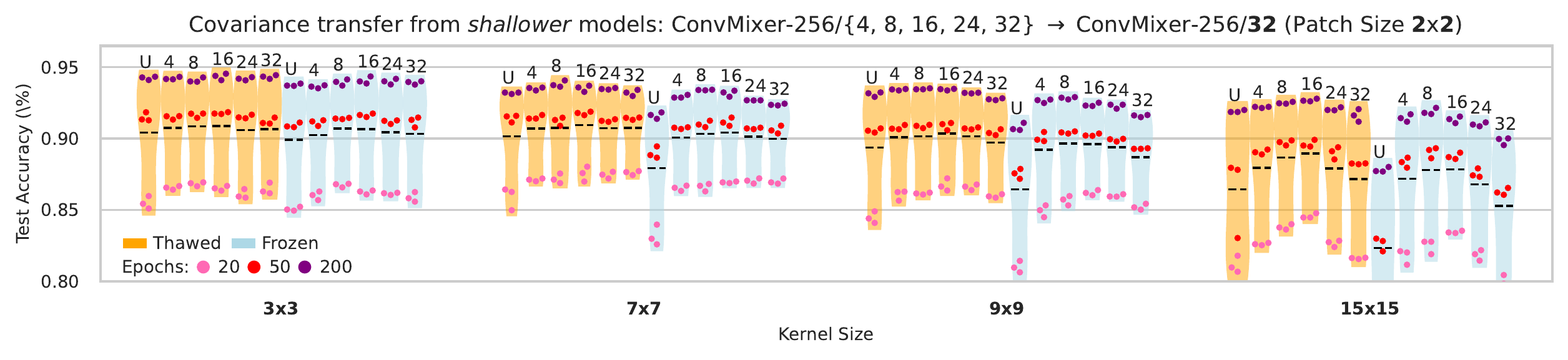}}
    \vspace*{-1.5em}
    \caption{CIFAR-10 experimental results from initializing via convariances from narrower \emph{(top)} and shallower \emph{(bottom)} models.
    The numeric annotations represent the width \emph{(top)} and depth \emph{(bottom)} of the pre-trained model we use to
    intialize. \textbf{U} represents uniform initialization.}
	\label{fig:width-transfer}
\end{figure}

\paragraph{Narrower models.} We first see if it's possible to train a narrower reference
model to calculate filter covariances to initialize a wider model;
for example, using a ConvMixer-32/8 to initialize a ConvMixer-256/8.
In Figure~\ref{fig:width-transfer}, we show that the optimal
performance surprisingly \emph{comes from the covariances of a smaller model}.
For filter sizes sizes greater than 3, 
the covariance transfer performance increases with width until width 32,
and then decreases for width 256 for both the thawed and frozen cases.
We plot this method in Fig.~\ref{fig:cifar-minit} (group \textbf{C}),
and note that it almost uniformly exceeds the performance of covariance transfer
from the same-sized model.
Note that the method does not change; the covariances are simply calculated from a smaller sample of filters.

\paragraph{Shallower models.}
\looseness=-1
Covariance transfer from a shallow model to a deeper model is somewhat
more complicated, as there is no longer a one-to-one mapping between layers.
Instead, we \emph{linearly interpolate} the covariance matrices to the desired depth.
Surprisingly, we find that this technique is also highly effective: for example,
for a 32-layer-deep ConvMixer, the optimal covariance transfer result is from an 8-layer-deep ConvMixer, and 2- and 4-deep
models are also quite effective (see Figure~\ref{fig:width-transfer}).

\paragraph{Different patch sizes.}
Similarly, it is straightforward to transfer covariances between models with different patch sizes.
We find that initializing ConvMixers with $1\times 1$ patches from filter covariances of ConvMixers with  $2\times 2$ patches
leads to a decrease in performance relative to using a reference model of the correct patch size;
however, using the filters of a $1 \times 1$ patch size ConvMixer to initialize a $2\times2$ patch size ConvMixer
increases performance (see group \textbf{b} vs. group \textbf{B} in Fig.~\ref{fig:patch-transfer}).
Yet, in both cases, the performance is better than uniform initialization.

\paragraph{Different filter sizes.}
Covariance transfer between models with different filter sizes is more challenging,
as the covariance matrices have different sizes.
In the block form, we mean-pad or clip each block to the target filter size, and then bilinearly interpolate over the blocks
to reach a correctly-sized covariance matrix.
This technique is still better than uniform initialization for filter sizes larger than 3
(which naturally has very little structure to transfer),
especially in the frozen case (see Fig. \ref{fig:patch-transfer})

\paragraph{Discussion.}
We have demonstrated that it is possible to initialize filters from the covariances
of pre-trained models of different widths, depths, patch sizes, and kernel sizes;
while some of these techniques perform better than others, they are almost all better than uniform
initialization.
Our observations indicate that the optimal choice of reference model
is narrower or shallower, and perhaps with a smaller patch size or kernel size.
We also found that covariance transfer from ConvMixers trained on ImageNet led
to greater performance still~(Appendix \ref{apx:cifar}).
This suggests that the best covariances for filter initialization
may be quite unrelated to the target model, \emph{i.e.,} model independent.

\section{D.I.Y. Filter Covariances}

Ultimately, the above methods for initializing convolutional filters via transfer are limited by the necessity of a trained network from which to form a filter distribution, which must be accessible at initialization.
We thus use observations on the structure of filter covariance matrices to construct our own covariance matrices from scratch.
Using our construction, we propose a depth-dependent but simple initialization
strategy for convolutional filters that greatly outperforms previous techniques.

\paragraph{Visual observations.}
Filter covariance matrices in pre-trained ConvMixers and ConvNeXts
have a great deal of structure, which we observe across models with different patch sizes,
architectures, and data sets; see Fig.~\ref{fig:cifar-p1-covs} and \ref{fig:imnet-covs}
for examples.
In both the block and rearranged forms of the covariance matrices,
we noticed clear repetitive structure,
which led to an initial investigation on modeling
covariances via Kronecker factorizations;
see Appendix~\ref{apx:cifar} for experimental results.
Beyond this, we first note that the overall variance of filters tends
to increase with depth, until breaking down towards the last layer.
Second, we note that the sub-blocks of the covariances
often have a \emph{static} negative component in the center,
with a \emph{dynamic} positive component whose position
mirrors that of the block itself.
Finally, the covariance of filter parameters is greater
in their center, \emph{i.e.,} covariance matrices are at first
centrally-focused and become more diffuse with depth.
These observations agree with intuition about the structure of convolutional filters:
most filters have the greatest weight towards their center,
and their parameters are correlated with their neighbors.

\paragraph{Constructing covariances.}
With these observations in mind, we propose a construction of covariance matrices.
We fix the (odd) filter size
$k \in \mathbb{N}^+$, let $\mathbf{1} \in \mathbb{R}^{k \times k}$ be the all-ones matrix,
and, as a building block for our initialization, use unnormalized Gaussian-like
filters $Z_\sigma \in \mathbb{R}^{k \times k}$ with a single variance parameter $\sigma$,
defined elementwise by
\begin{equation}
    (Z_\sigma)_{i,j} := \exp\left(-\frac{(i-\lfloor\frac{k}{2}\rfloor)^2 + (j-\lfloor\frac{k}{2}\rfloor)^2}{2\sigma}\right)\text{    for }1 \leq i, j, \leq k.
\label{eq:gaussian-def}
\end{equation}
Such a construction produces filters similar to those observed in the blocks of the
Layer \#5 covariance matrix in Fig.~\ref{fig:cifar-p1-covs}.

To capture the \emph{dynamic} component that moves according to the position
of its block, we define the block matrix $C \in \mathbb{R}^{k^2\times k^2}$
with $k\times k$ blocks by
\begin{equation}
    [C_{i,j}] = \text{Shift}(Z_{\sigma}, i   - \lfloor\tfrac{k}{2}\rfloor , j   - \lfloor\tfrac{k}{2}\rfloor     )
\label{eq:shift}
\end{equation}
where the Shift operation translates each element of the matrix $i$ and $j$ positions forward in their respective dimensions, wrapping around when elements overflow; see Appendix~\ref{apx:shift-proof} for details.
We then define two additional components, both constructed from Gaussian filters: a \textit{static} component $S = \mathbf{1} \otimes Z_\sigma \in \mathbb{R}^{k^2 \times k^2}$ and a blockwise mask component $M = Z_\sigma \otimes \mathbf{1} \in \mathbb{R}^{k^2 \times k^2}$, which encodes higher variance as pixels approach the the center of the filter.

\begin{figure}[t]
	$$
        \overset{\Sigma}{\includegraphics[height=1.5cm,valign=c]{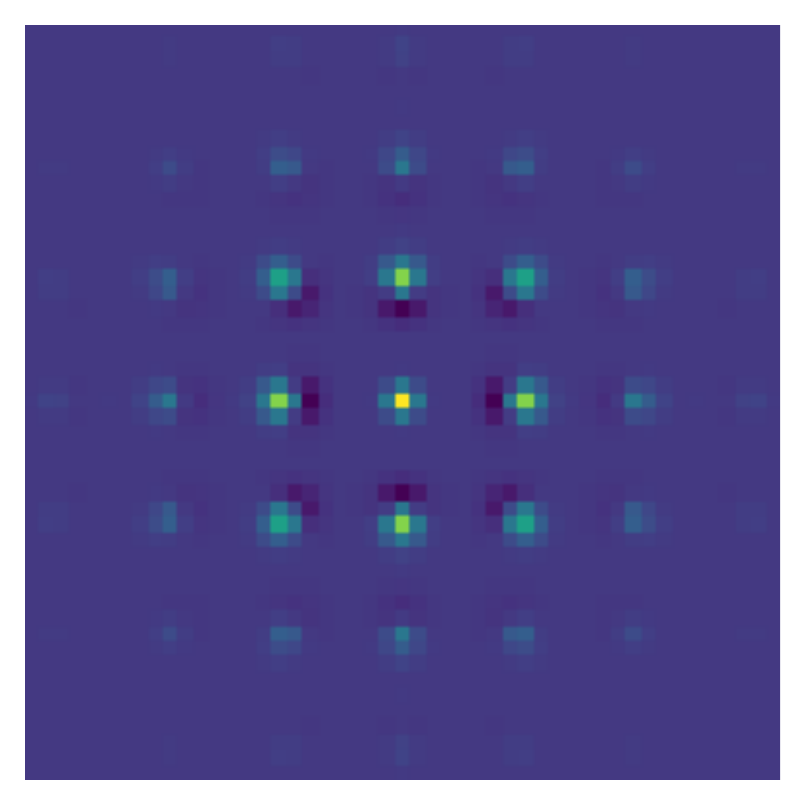}}
        \;
        =
        \;
        \overset{M}{\includegraphics[height=1.5cm,valign=c]{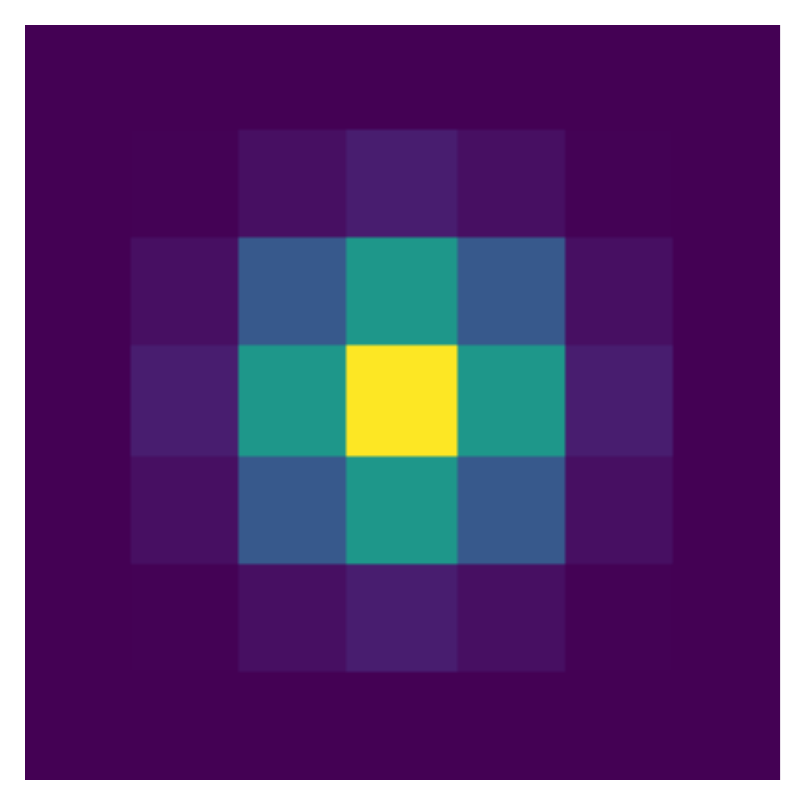}}
        \;
        \odot
        \;
        \Bigg(
        \;
        \overset{C}{\includegraphics[height=1.5cm,valign=c]{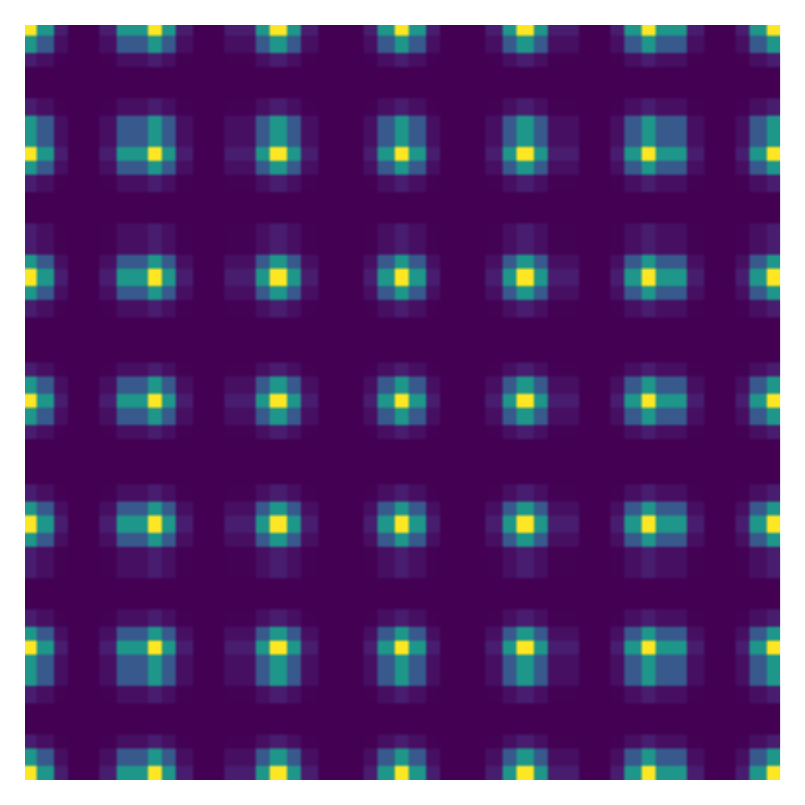}}
        \;
        -
        \;
        \overset{S}{\includegraphics[height=1.5cm,valign=c]{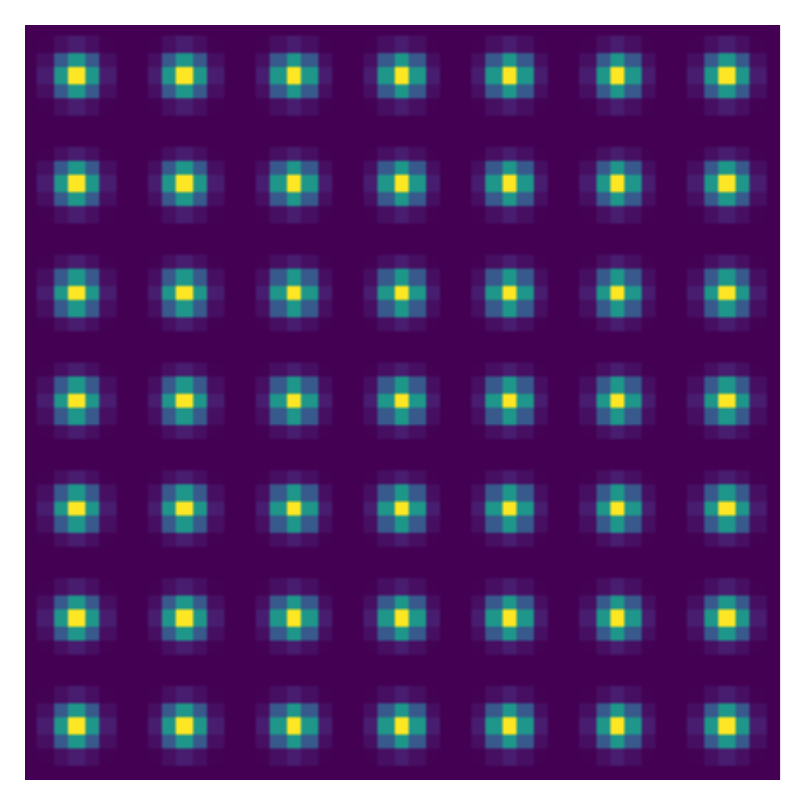}}
        \;
        \Bigg)
        \;
        +
        \;
        \overset{C}{\includegraphics[height=1.5cm,valign=c]{notebooks/gfx/C.pdf}}
        \;
        \odot
        \;
        \overset{S}{\includegraphics[height=1.5cm,valign=c]{notebooks/gfx/S.pdf}}
	$$
	\center
    \caption{Our convolutional covariance matrix construction with $\sigma = \pi/2$.}
    \label{fig:parameterization}
\end{figure}

Using these components and our intuition, we first consider
$\hat{\Sigma} = M \odot (C - \tfrac{1}{2}S)$, where $\odot$ is an elementwise
product. While this adequately represents what we view to be the important
structural components of filter covariance matrices,
it does not satisfy the property 
$\left[ \Sigma_{i,j}\right]_{\ell, m} = \left[ \Sigma_{\ell, m}\right]_{i,j}$
(\emph{i.e.,} covariance matrices must be symmetric, accounting for our block representation).
Consequently, we instead calculate its symmetric part, using
the notation as follows to denote a ``block-transpose'':
\begin{equation}
\Sigma^B = \Sigma' \Longleftrightarrow \left[ \Sigma_{i,j}\right]_{\ell, m} = \left[ \Sigma^\prime_{\ell, m}\right]_{i,j}  \text{ for } 1 \leq i, j, \ell, m \leq k.
\end{equation}
Equivalently, this is the perfect shuffle permutation such that $(X \otimes Y)^B = Y \otimes X$ with $X, Y \in \mathbb{R}^{k\times k}$.
First, we note that $C^B = C$ due to the definition of the shift operation used in Eq.~\ref{eq:shift} (see Appendix~\ref{apx:shift-proof}).
Then, noting that $S^B = M$ and $M^B = S$ by the previous rule, we define our construction of $\Sigma$ to be the symmetric part
of $\hat{\Sigma}$ (where $C, S, M$ are implicitly parameterized by $\sigma$, similarly to $Z_\sigma$):
\begin{align}
    \Sigma &= \tfrac{1}{2}(\hat{\Sigma} + \hat{\Sigma}^B)  = \tfrac{1}{2} \left[ M \odot (C - \tfrac{1}{2} S ) + (M \odot (C - \tfrac{1}{2} S))^B\right] \\
    &=   \tfrac{1}{2} \left[ M \odot (C - \tfrac{1}{2} S ) + (M^B \odot (C^B - \tfrac{1}{2} S^B))\right] =   M \odot (C - \tfrac{1}{2} S) + S \odot ( C - \tfrac{1}{2} M) \\
    &= \tfrac{1}{2} \left[M \odot (C - S) + S \odot C\right].
\end{align}
While $\Sigma$ is now symmetric (in the rearranged form of Eq.~\ref{eq:rearr}),
it is not positive semi-definite, but can easily be projected to $\mathbb{S}_+^{k^2}$, as is often done
automatically by multivariate Gaussian procedures.
We illustrate our construction in Fig.~\ref{fig:parameterization},
and provide an implementation in Fig.~\ref{fig:code}.

\begin{figure}
  \begin{minipage}[c]{0.65\textwidth}
    \includegraphics[width=\textwidth,trim=10 150 10 35,clip]{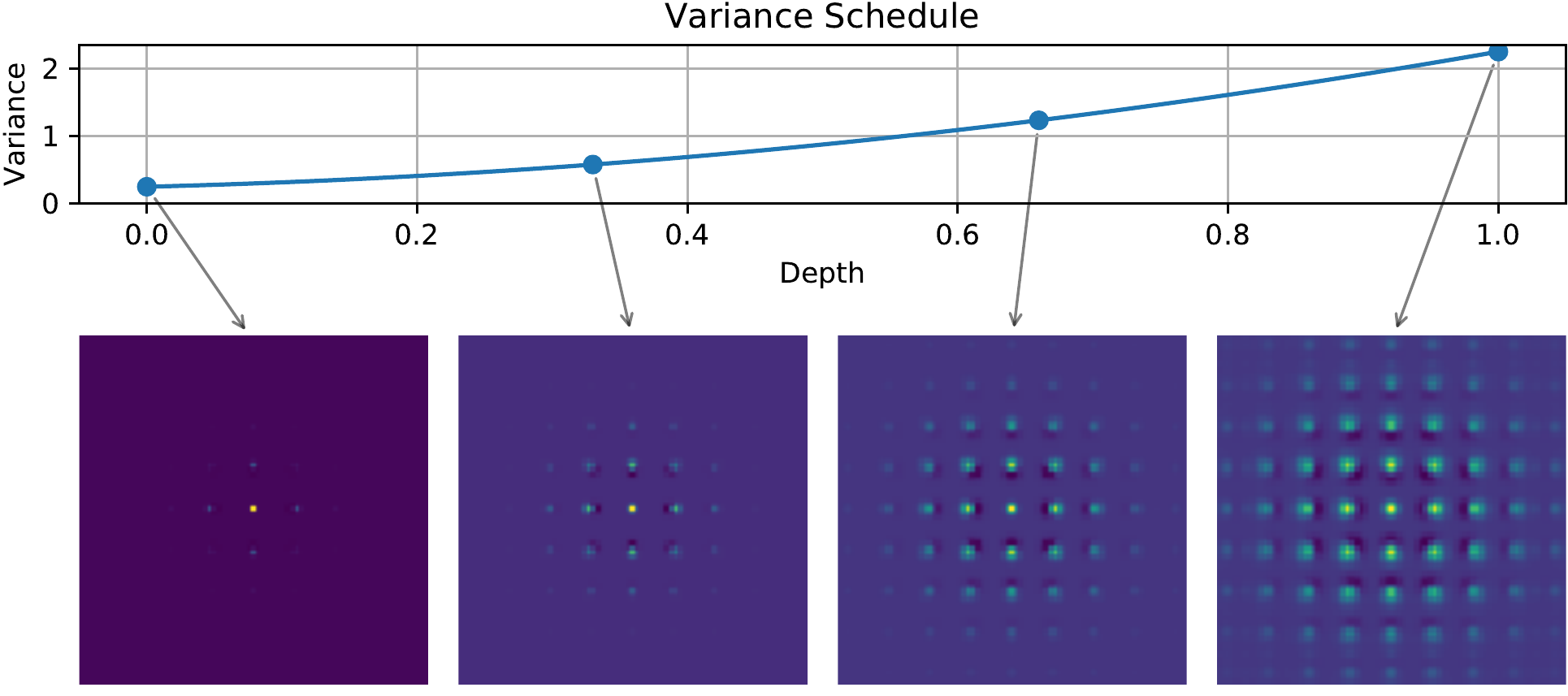}
  \end{minipage}\hfill
  \begin{minipage}[c]{0.32\textwidth}
    \caption{How our initialization
        changes with depth.
        Variance increases quadratically with depth
        according to a schedule
        which can be chosen
        through visual inspection
        of pre-trained models
        or through grid search.
        Here we use the parameters \newline
        $\sigma_0 = .5$, $v_\sigma=.5$, $a_\sigma = 3$.
    } \label{fig:varsched}
  \end{minipage}
\end{figure}

\paragraph{Completing the initialization.}
As explained in Fig.~\ref{fig:cifar-p1-covs},
we observed that in pre-trained models, the filters become more ``diffuse''
as depth increases; we capture this fact in our construction
by increasing the parameter $\sigma$ with depth according to a simple quadratic schedule;
let $d$ be the percentage depth, i.e., $d = \tfrac{i-1}{D-1}$ for the $i^{\text{th}}$
convolutional layer of a model with $D$ total such layers. Then for layer $i$, we
parameterize our covariance construction by a \emph{variance schedule}:
\begin{equation}
    \sigma(d) = \sigma_0 + v_\sigma d + \tfrac{1}{2} a_\sigma d^2
\end{equation}
where $\sigma_0, v_\sigma, a_\sigma$ jointly describe how the covariance evolves with depth. 
Then, for each layer $i \in 1, \dots, D$, we compute $d = \tfrac{i-1}{D-1}$
and initialize the filters as $F_{i,j} \sim \mathcal{N}(0, \Sigma'_{\sigma(d)})$ for $j \in 1, \dots, H$.
We illustrate our complete initialization scheme in Figure~\ref{fig:varsched}.

\section{Results}

In this section, we present the performance of our initialization
within ConvMixer and ConvNeXt on CIFAR-10 and ImageNet classification,
finding it to be highly effective, particularly for deep models with large filters.
Our new initialization overshadows our previous covariance transfer results.

\looseness=-1
Settings of initialization hyperparameters $\sigma_0$, $v_\sigma$, and $a_\sigma$ were found and fixed 
for CIFAR-10 experiments, while two such settings were used for ImageNet experiments.
Appendix \ref{sec:hyper} contains full details on our (relatively small) hyperparameter searches 
and experimental setups, as well as empirical evidence that our method is \emph{robust to a large swath of hyperparameter settings.}

\subsection{CIFAR-10 Results}

\begin{wrapfigure}[10]{r}{0.5\textwidth}
 \vspace*{-1em}
  \begin{center}
    \includegraphics[height=2.9cm,trim=7 5 10 5,clip]{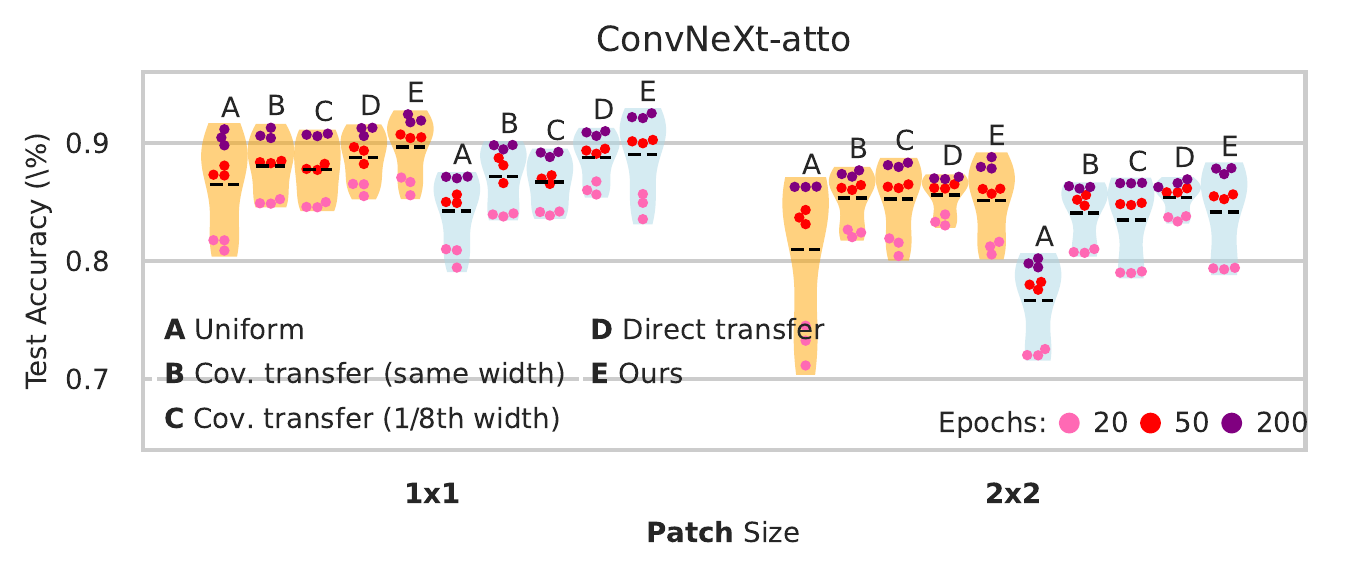}
  \end{center}
	\vspace*{-1.25em}
    \caption{Our init also improves ConvNeXt's accuracy on CIFAR-10 (group \textbf{E} \emph{vs.} \textbf{A}).}
  \label{fig:minit-convnext}
\end{wrapfigure}

\paragraph{Thawed filters.}
\looseness=-1
In Fig.~\ref{fig:cifar-minit},
we show that large-kernel models using our initialization
(group \textbf{E})
outperform those using uniform initialization (group \textbf{A}),
covariance transfer (groups \textbf{B, C}),
and even those directly initializing via
learned filters (group \textbf{D}).
For $2\times 2$-patch models (200 epochs), relative to uniform,
our initialization causes up to a 
1.1\% increase in accuracy for ConvMixer-256/8,
and up to 1.6\% for ConvMixer-256/24.
The effect size increases with the the filter size,
and is often more prominent for shorter training times.
Results are similar for $1\times 1$-patch models,
but with a smaller increase for $7\times 7$ filters
($0.15\%$ vs. $0.5\%$).
Our initialization has the same effects for ConvNeXt (Fig.~\ref{fig:minit-convnext}).
However, our method works poorly for $3\times 3$ filters,
which we believe have fundamentally different structure than larger filters;
this setting is better-served by our original covariance transfer techniques.

\begin{figure}[h!]
	\centering
	\subfigure{\includegraphics[width=0.47\textwidth]{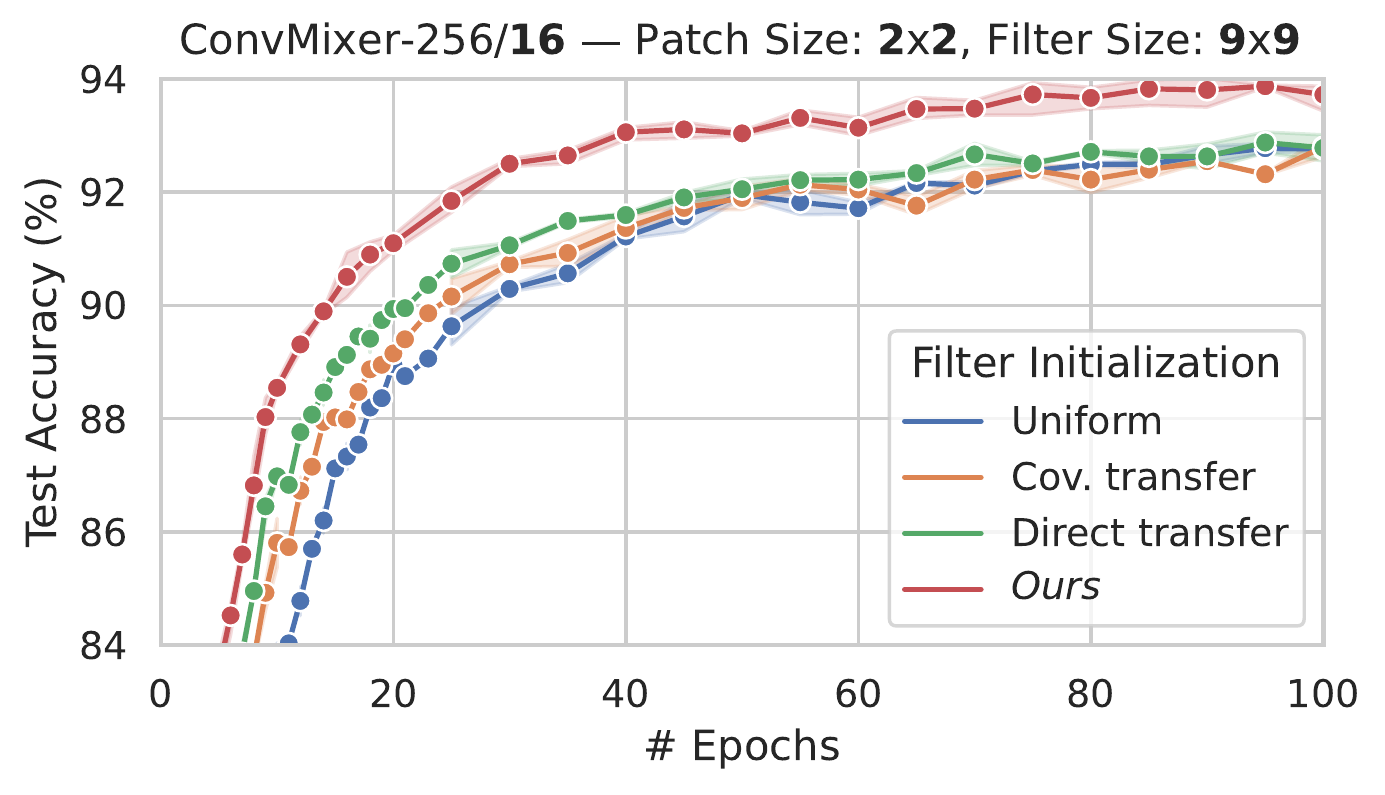}}
	\subfigure{\includegraphics[width=0.47\textwidth]{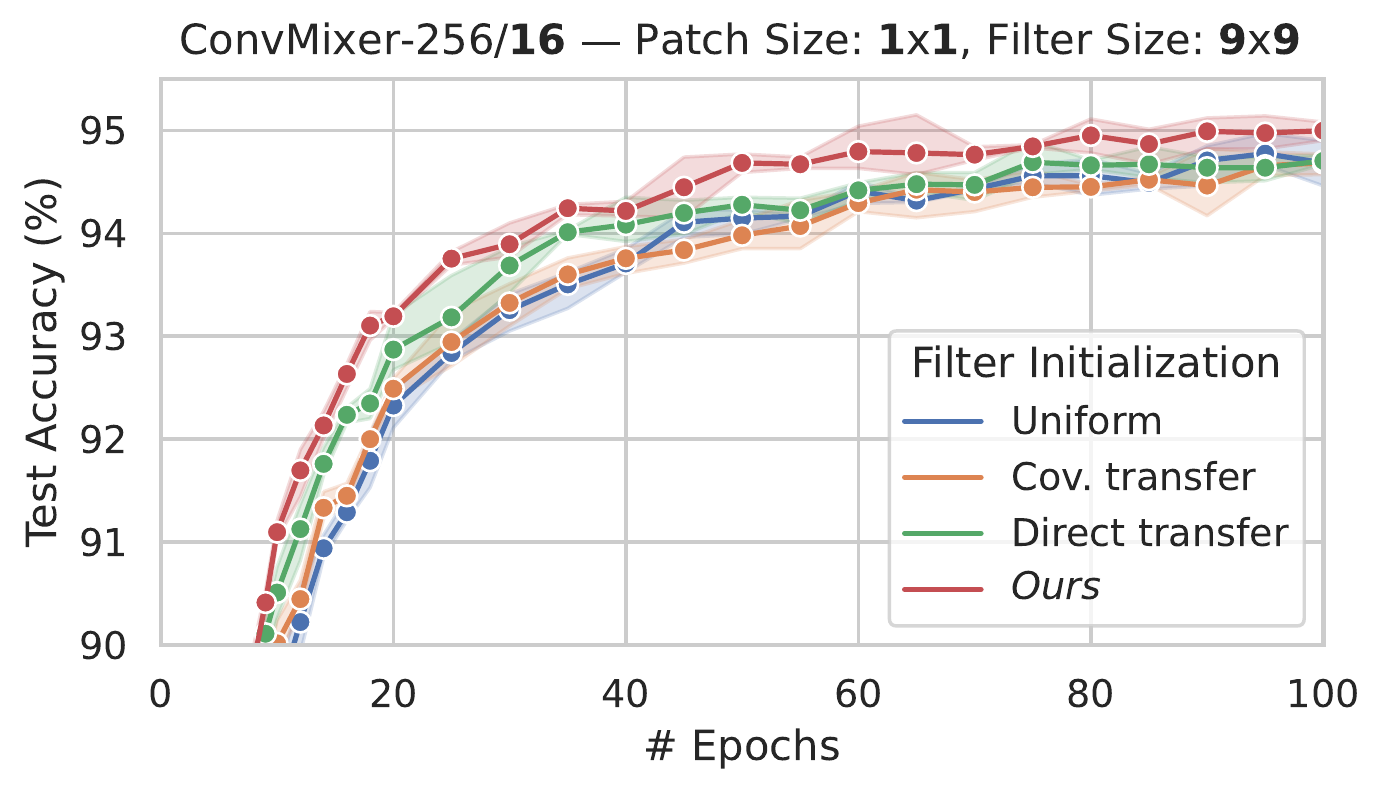}}
    \vspace*{-1em}
	\caption{Convergence plots: each data point runs through a full cycle of the LR schedule,
    and all points are averaged over three trials with shaded standard deviation.}
	\label{fig:minit-converge}
\end{figure}

In addition to improving the final accuracy,
our initialization also drastically speeds up convergence
of models with thawed filters
(see Fig.~\ref{fig:minit-converge}), particularly for deeper models.
A ConvMixer-256/16 with $2\times 2$ patches using our initialization reaches
$90\%$ accuracy in approximately $50\%$ fewer epochs
than uniform initialization,
and around $25\%$ fewer than direct learned filter transfer.
The same occurs, albeit to a lesser extent, for $1\times1$ patches%
---but note that for this experiment we used the same initialization parameters for both patch sizes
to demonstrate robustness to parameter choices.

\paragraph{Frozen filters.}
\looseness=-1
Our initialization leads to
even more surprising effects in models with frozen filters.
In Fig.~\ref{fig:cifar-minit},
we see that frozen-filter $2\times 2$-patch models
using our initialization
often \emph{exceed the performance of their uniform, thawed-filter counterparts}
by a significant margin of 0.4\% -- 2.0\% for 200 epochs,
and an even larger margin of 0.6\% -- 5.0\% for 20 epochs
(for large filters).
That is, group \textbf{E} \emph{(frozen})
consistently outperforms groups \textbf{A-D} \emph{(thawed)},
and in some cases even group \textbf{E} \emph{(thawed)}, % TODO might want to remove
especially for the deeper 24-layer ConvMixer.
While this effect breaks down for $1\times 1$ patch models,
such frozen-filter models still see accuracy increases of  0.6\%--3.5\%.
However, the effect can still be seen for $1\times 1$-patch ConvNeXts (Fig.~\ref{fig:minit-convnext}).
Also note that frozen-filter models can be up to 40\% faster to train
(see Fig.~\ref{fig:fast-freeze}), and may be more robust~\citep{rethinking}.

\begin{table}[t]
    \caption{ImageNet-1k accuracy from various architectures and initializations. ``Ours'' denotes our proposed initialization. \textbf{Bold} indicates best within architecture and category (frozen or thawed).}
    \label{tab:imnet}
    \vspace{0.5em}

\resizebox{\columnwidth}{!}{%
    \small
\begin{tabular}{|l|c|c|c|b|b|b|a|a|a|}
    \hline
    \multicolumn{4}{|c|}{\thead{Model}} & \multicolumn{3}{b|}{\textsc{Thawed}} & \multicolumn{3}{a|}{\textsc{Frozen}} \\
    \hline
    Architecture & \makecell{ {\scriptsize Filter} \\ {\scriptsize Size}}  & \makecell{ {\scriptsize Patch} \\ {\scriptsize Size}}  & \makecell{ {\scriptsize \#} \\ {\scriptsize Epochs} } & {\scriptsize Uniform} & \makecell{{\scriptsize Ours} \\ {\scriptsize .15 .5 .25} } & \makecell{ {\scriptsize Ours} \\ {\scriptsize .15 .25 1.0} } & {\scriptsize Uniform} & \makecell{ {\scriptsize Ours} \\ {\scriptsize .15 .5 .25} } & \makecell{{\scriptsize Ours} \\ {\scriptsize .15 .25 1.0}} \\
    \hline
    ConvMixer-512/12 & 9 & 14 & 50 &              67.03 & \textbf{67.41} & 67.34 &     60.47 & \textbf{64.43} & 64.12 \\
    ConvMixer-512/24 & 9 & 14 & 50 &              67.76 & \textbf{69.60} & 69.52 &              62.50 & \textbf{66.57} & 66.38 \\
    ConvMixer-512/32 & 9 & 14 & 50 &              65.00 & 68.78 & \textbf{68.84} &              55.79 & \textbf{66.59} & 66.32 \\
    ConvMixer-1024/12 & 9 & 14 & 50 &             73.55 & 73.62 & \textbf{73.75} &              68.96 & \textbf{71.48} & 71.30 \\
    ConvMixer-1024/24 & 9 & 14 & 50 &             74.19 & 75.33 & \textbf{75.50} &              69.65 & \textbf{73.42} & 74.31 \\
    ConvMixer-1024/32 & 9 & 14 & 50 &             72.18 & \textbf{74.98} & 74.95 &              64.94 & 73.00 & \textbf{73.12} \\
    ConvMixer-512/12 & 9 & 7 & 50 &               72.05 & 71.92 & \textbf{72.32} &              67.25 & 68.91 & \textbf{68.92} \\
    \hline
    ConvNeXt-Atto & 7 & 4 & 50 &                  \textbf{69.96} & 67.84 & 68.06 &              51.43 & \textbf{64.52} & 64.43 \\
    ConvNeXt-Tiny & 7 & 4 & 50 &                  75.99 & 76.08 & \textbf{77.11} &              64.17 & 74.62 & \textbf{75.21} \\
    \hline
    ConvMixer-1536/24 & 9 & 14 & 150 &            80.11 & \multicolumn{1}{g|}{}& \textbf{80.28} & \multicolumn{3}{g|}{}          \\
    ConvNeXt-Tiny & 7 & 4 & 150 &                 79.74 & \multicolumn{1}{g|}{}& \textbf{79.81} & \multicolumn{3}{g|}{}          \\
    \hline
\end{tabular}
    }
\end{table}

\subsection{ImageNet Experiments}
Our initialization performs extremely well on CIFAR-10 for large-kernel models,
almost always helping and rarely hurting.
Here, we explore if the performance gains transfer to larger-scale ImageNet models.
We observe in  Fig. \ref{fig:imnet-covs}, Appendix~\ref{apx:imnet-exp}
that filter covariances for such models have finer-grained structure
than models trained on CIFAR-10, perhaps due to using larger patches.
Nonetheless, our initialization leads to quite encouraging improvements
in this setting.

\paragraph{Experiment design.}
We used the ``A1'' training recipe from~\cite{ross},
with cross-entropy loss, fewer epochs, and a triangular LR schedule
as in~\cite{trockman}.
We primarily demonstrate our initialization
for 50-epoch training, as the difference
between initializations is most pronounced for
lower training times.
We also present two full, practical-scale 150-epoch experiments
on large models.
We also included covariance transfer experiments
in Appendix~\ref{apx:imnet-exp}.

\paragraph{Thawed filters.}
On models trained for 50 epochs with thawed filters,
our initialization improves the final
accuracy by $0.4\% - 3.8\%$ (see Table~\ref{tab:imnet}).
For the relatively-shallow ConvMixer-512/12
on which we tuned the initialization parameters,
we see a gain of just $0.4\%$;
however, when increasing the depth to 24 or 32,
we see larger gains of $1.8\%$ and $3.8\%$, respectively, 
and a similar trend among the wider ConvMixer-1024 models.
Our initialization also boosts the accuracy
of the 18-layer ConvNeXt-Tiny from $76.0\%$ to $77.1\%$;
however, it decreased the accuracy of the smaller, 12-layer ConvNeXt-Atto.
This is perhaps unsurprising,
seeing as our initialization seems to be more helpful for deep models,
and we used hyperparameters optimized for a model with a substantially different
patch and filter size.

Our initialization is also beneficial for more-practical 150-epoch training,
boosting accuracy by around $0.1\%$ on both ConvMixer-1536/24 and ConvNeXt-Tiny
(see Table~\ref{tab:imnet}, bottom rows).
While the effect is small, this demonstrates that our initialization is still helpful 
even for longer training times and very wide models.
We expect that within deeper models and with slightly more parameter tuning,
our initialization could lead to still larger gains in full-scale ImageNet training.

\paragraph{Frozen filters.}
Our initialization is extremely helpful
for models with frozen filters.
Using our initialization, the difference between thawed
and frozen-filter models decreases with increasing depth,
\emph{i.e.,} it leads to $2- 11\%$ improvements
over models with frozen, uniformly-initialized filters.
For ConvMixer-1024/32, the accuracy improves from $64.9\%$
to $73.1\%$, which is over $1\%$ \emph{better than
the corresponding thawed, uniformly-initialized model},
and only $2\%$ from the best result using our initialization.
This mirrors the effects we saw for deeper models on our earlier CIFAR-10 experiments.
We see a similar effect for ConvNeXt-Tiny, with the frozen version
using our initialization achieving $75.2\%$ accuracy \emph{vs.}
the thawed $76.0\%$.
In other words, our initialization so effectively captures the structure
of convolutional filters that it is hardly necessary to train them after initialization;
one benefit of this is that it substantially speeds up training for large-filter convolutions.

\section{Conclusion}

In this paper, we proposed a simple, closed-form,
and learning-free initialization
scheme for large depthwise convolutional filters.
Models using our initialization typically reach higher accuracies
more quickly than uniformly-initialized models.
We also demonstrated that our random initialization of convolutional
filters is so effective, that in many cases, networks
perform nearly as well (or even better) if the resulting filters do
not receive gradient updates during training.
Moreover, like the standard uniform initializations
generally used in neural networks,
our technique merely
samples from a particular statistical distribution,
and it is thus almost completely computationally free.
\emph{\textbf{In summary,} our initialization
technique for the increasingly-popular large-kernel depthwise convolution operation 
almost always helps, rarely hurts, and is also free.}

\bibliography{main}
\bibliographystyle{iclr2023_conference}

\clearpage
\appendix

\section{Additional CIFAR Results}
\label{apx:cifar}

\begin{figure}[h!]
	\centering
    \subfigure{\includegraphics[height=2.65cm,trim=7 20 0 5,clip]{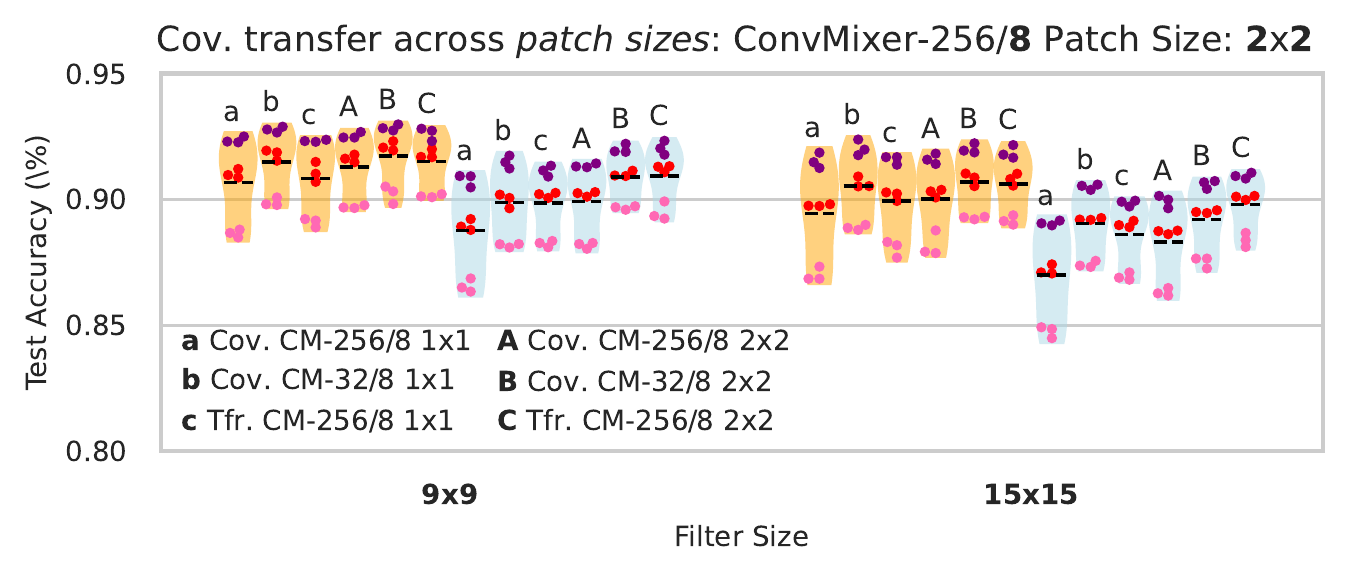}}
	\subfigure{\includegraphics[height=2.65cm,trim=39 20 0 5,clip]{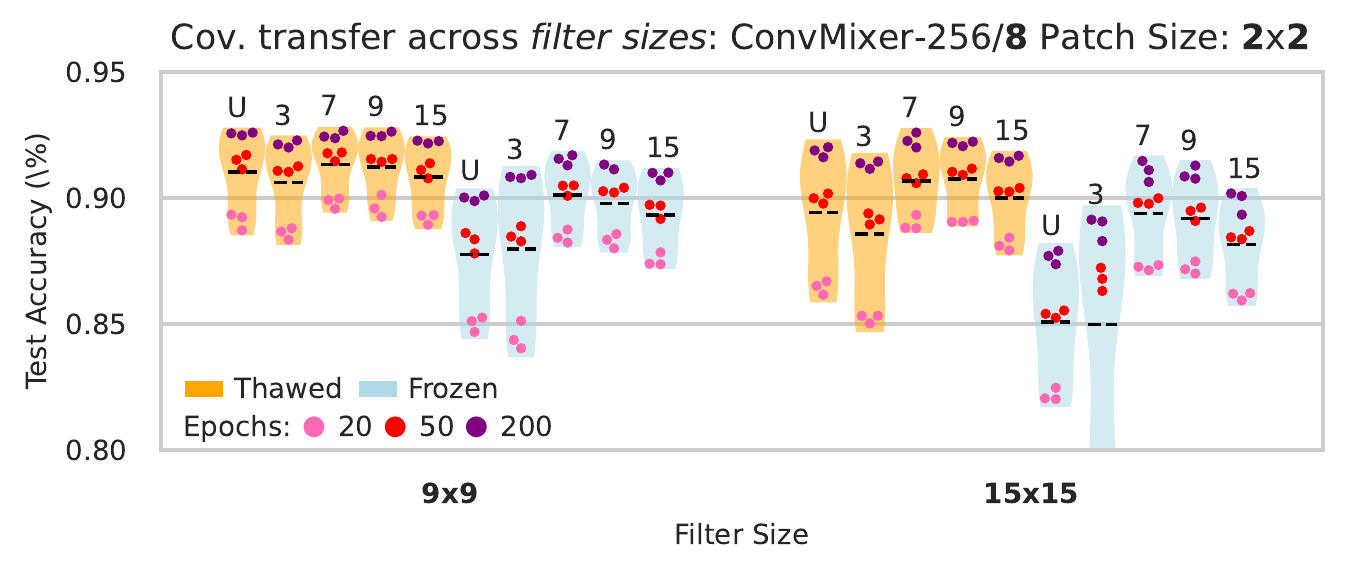}}
    \vspace*{-1em}
    \caption{Initializing via covariances from models with different patch (\emph{left}) and filter sizes (\emph{right}).
    \emph{Left:} Lowercase denotes initializing from patch size $1\times1$, and uppercase $2\times 2$.
    \emph{Right:} Annotations denote the reference filter size, \textbf{U} is uniform.}
	\label{fig:patch-transfer}
\end{figure}

\begin{figure}[h!]
	\centering
    \subfigure{\includegraphics[height=2.65cm,trim=7 20 0 5,clip]{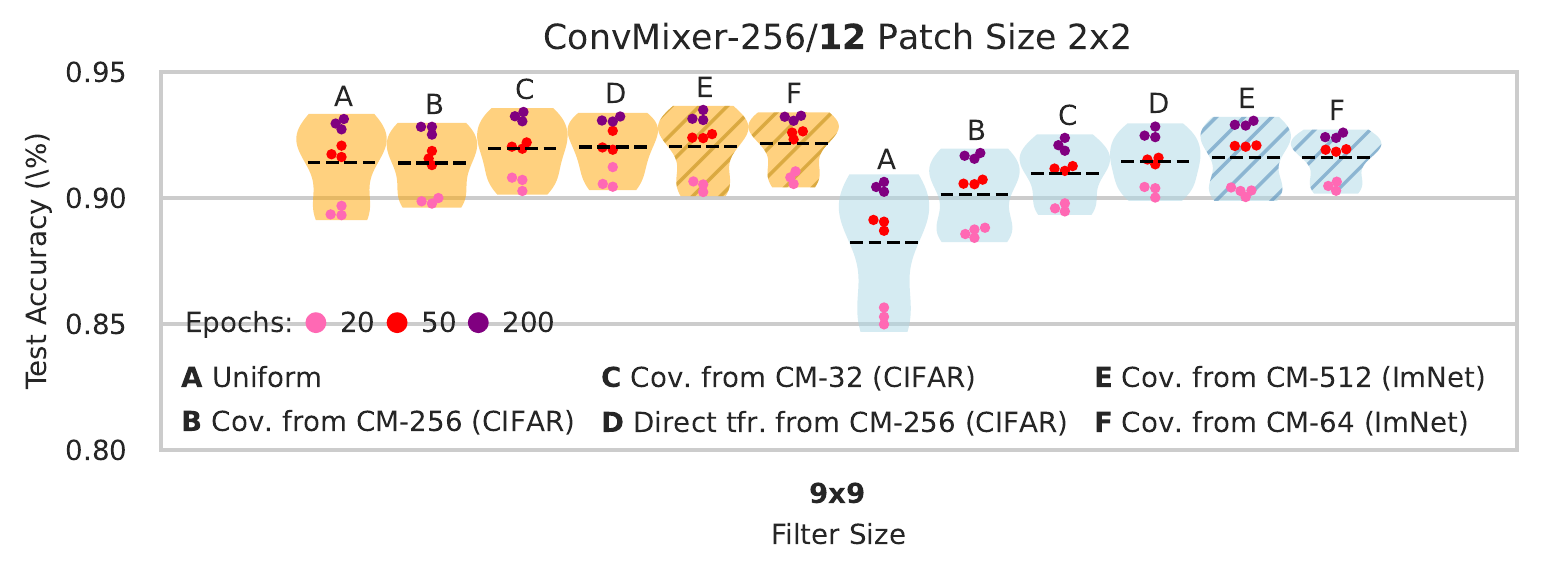}}
	\subfigure{\includegraphics[height=2.65cm,trim=7 20 0 5,clip]{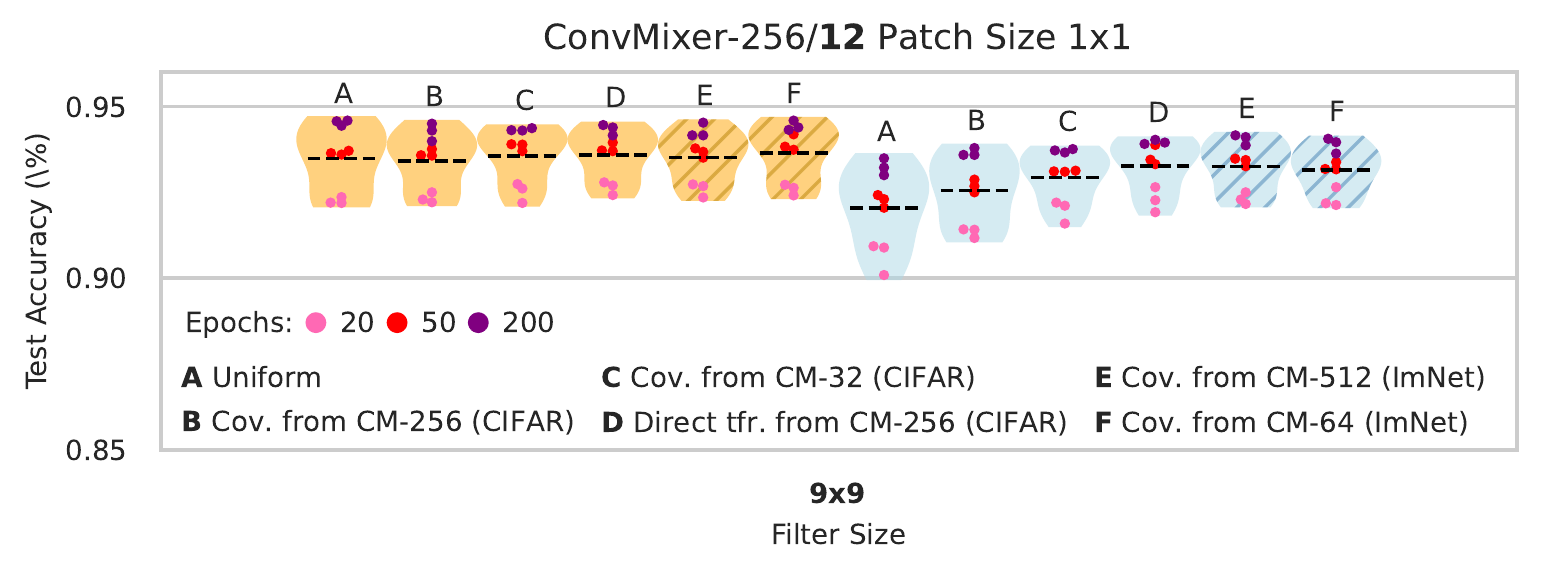}}
    \vspace*{-1em}
    \caption{
        Using filter distributions from pre-trained ImageNet models to initialize
        models trained on CIFAR-10 is also effective
        (represented by groups \textbf{E} and \textbf{F}, with hatch marks).
    }
	\label{fig:imnet-transfer}
\end{figure}

\begin{figure}[h!]
	\centering
    \subfigure{\includegraphics[height=2.65cm,trim=7 20 0 5,clip]{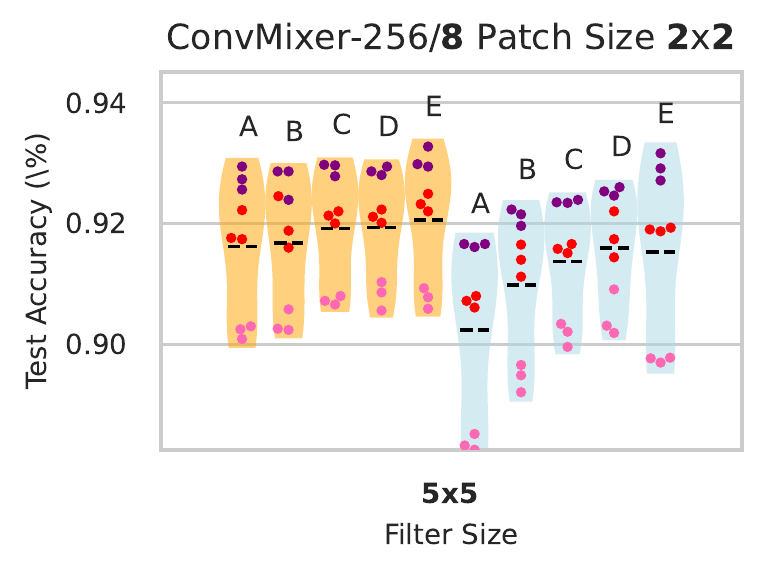}}
	\subfigure{\includegraphics[height=2.65cm,trim=18 20 0 5,clip]{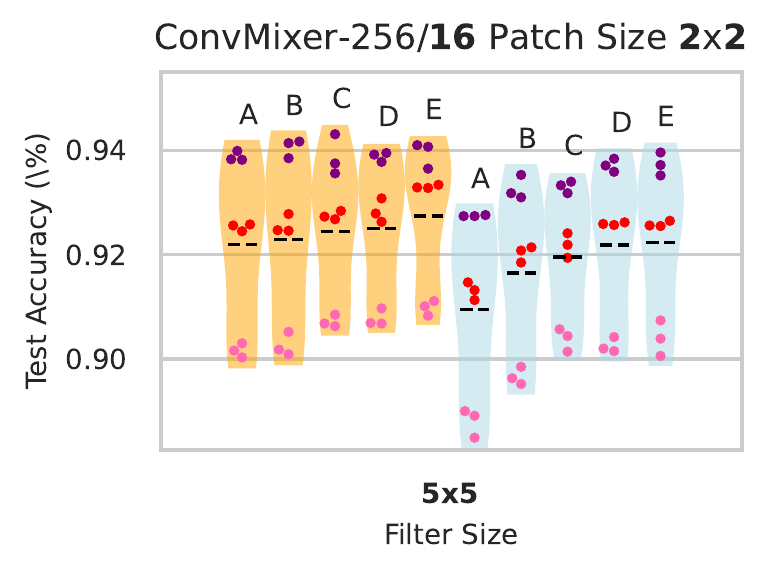}}
    \vspace*{-1em}
    \caption{
        Our initialization is also effective for $5\times 5$ filters.
        (The same legends in Fig.~\ref{fig:cifar-minit} apply.)
    }
	\label{fig:imnet-transfer}
\end{figure}

\begin{figure}[h!]
	\centering
    \subfigure{\includegraphics[width=0.47\textwidth]{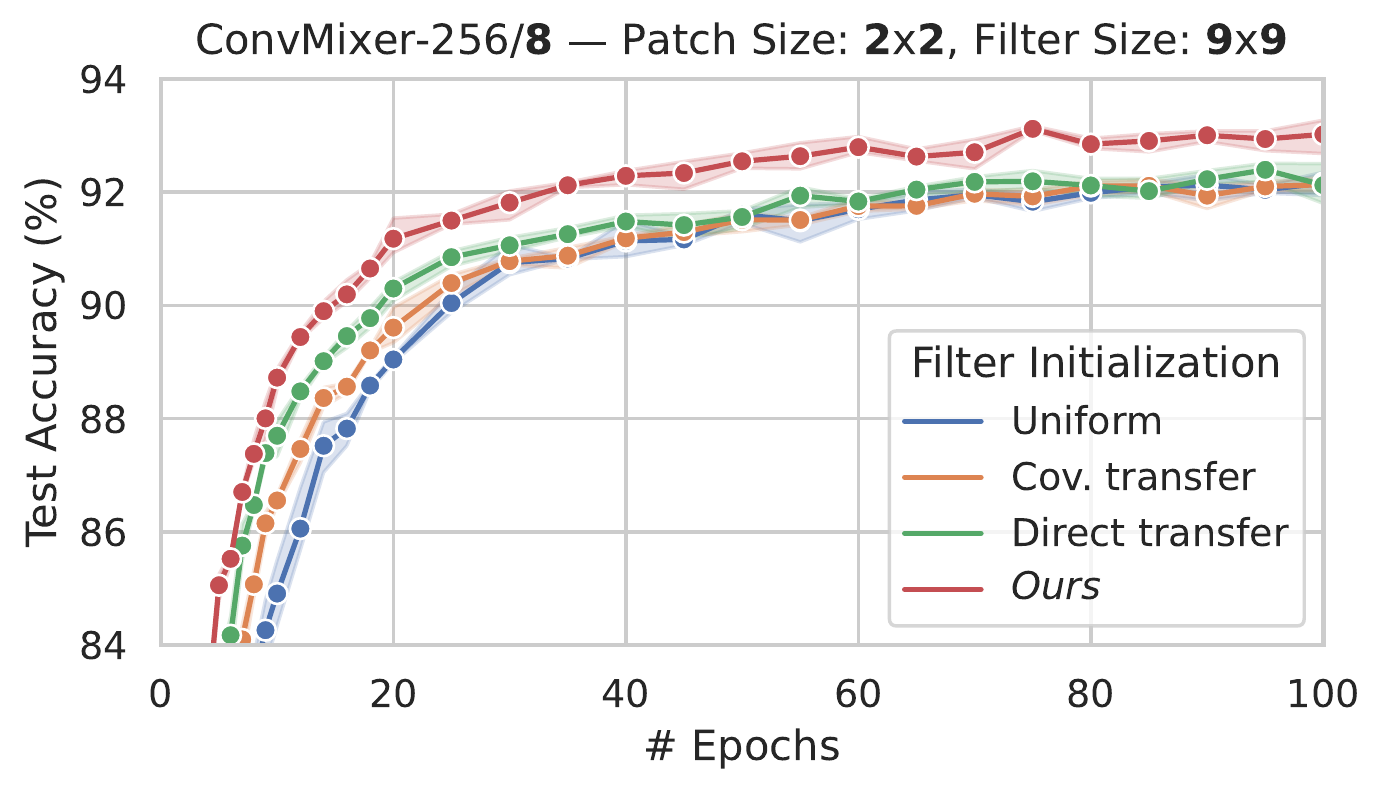}}
	\subfigure{\includegraphics[width=0.47\textwidth]{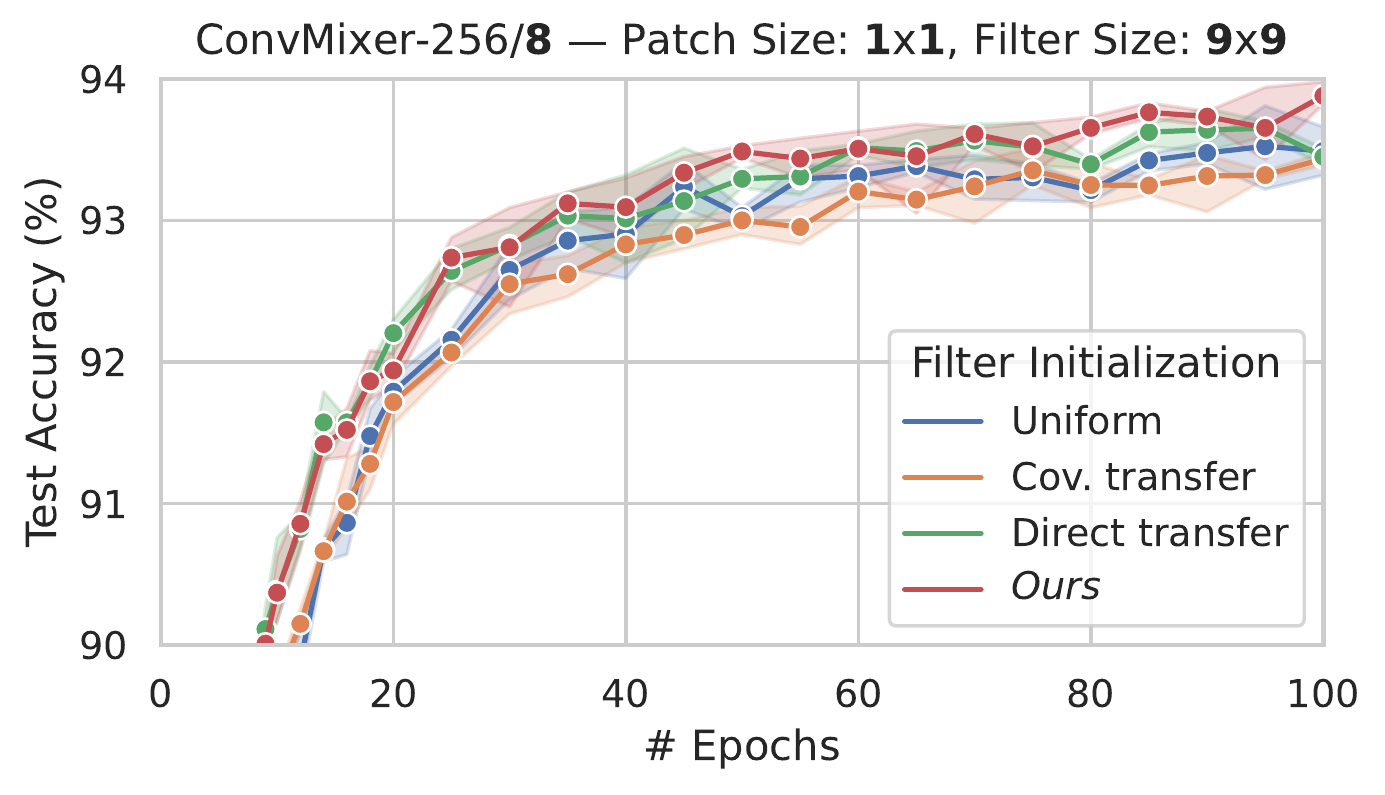}}
    \vspace*{-1em}
	\caption{Convergence plots: each data point runs through a full cycle of the LR schedule,
    and all points are averaged over three trials with shaded standard deviation.}
	\label{fig:minit-converge-apx}
\end{figure}

\clearpage

\begin{figure}[h!]
    \centering
    \subfigure{\includegraphics[width=0.7\textwidth]{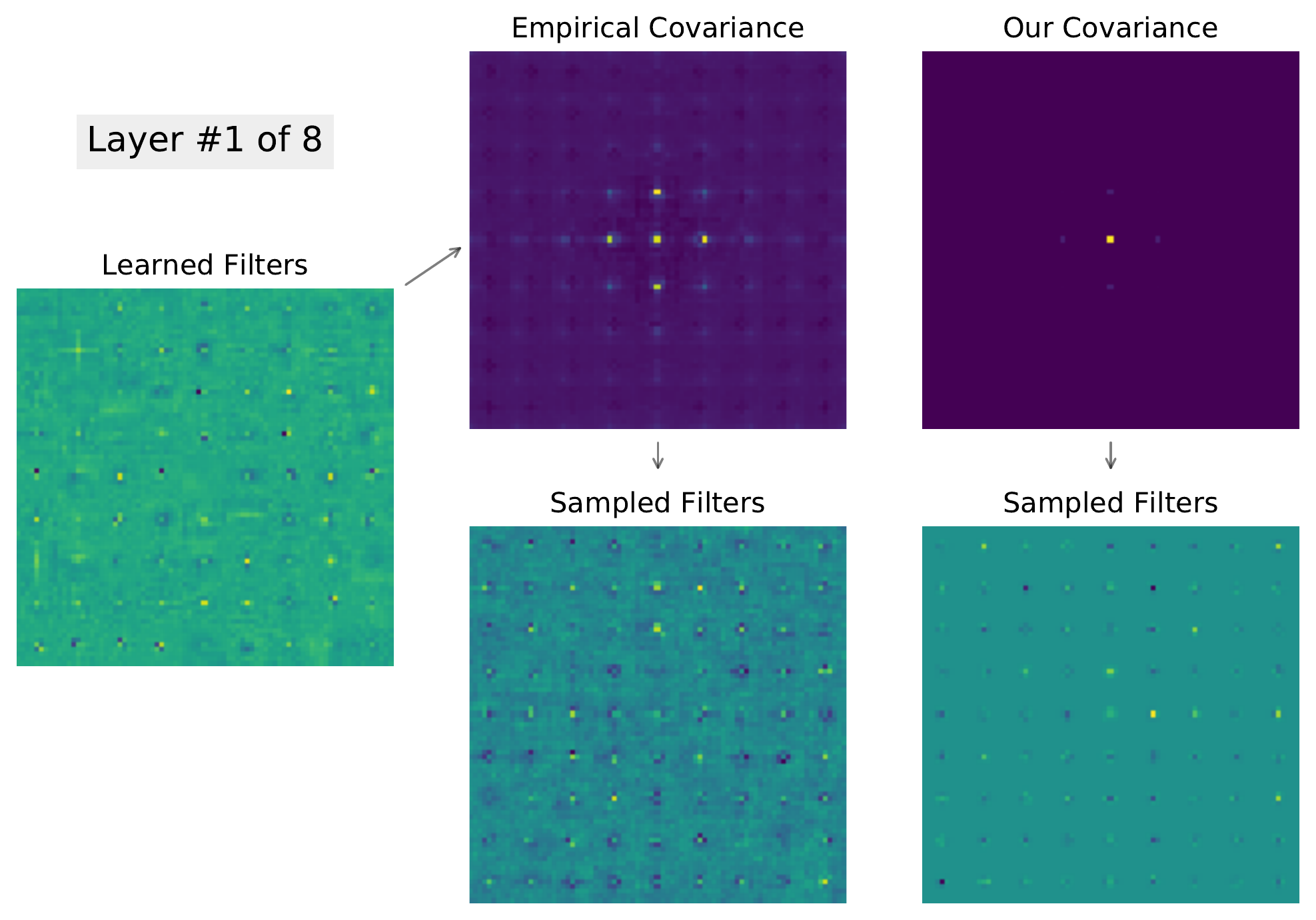}}
    \subfigure{\includegraphics[width=0.7\textwidth]{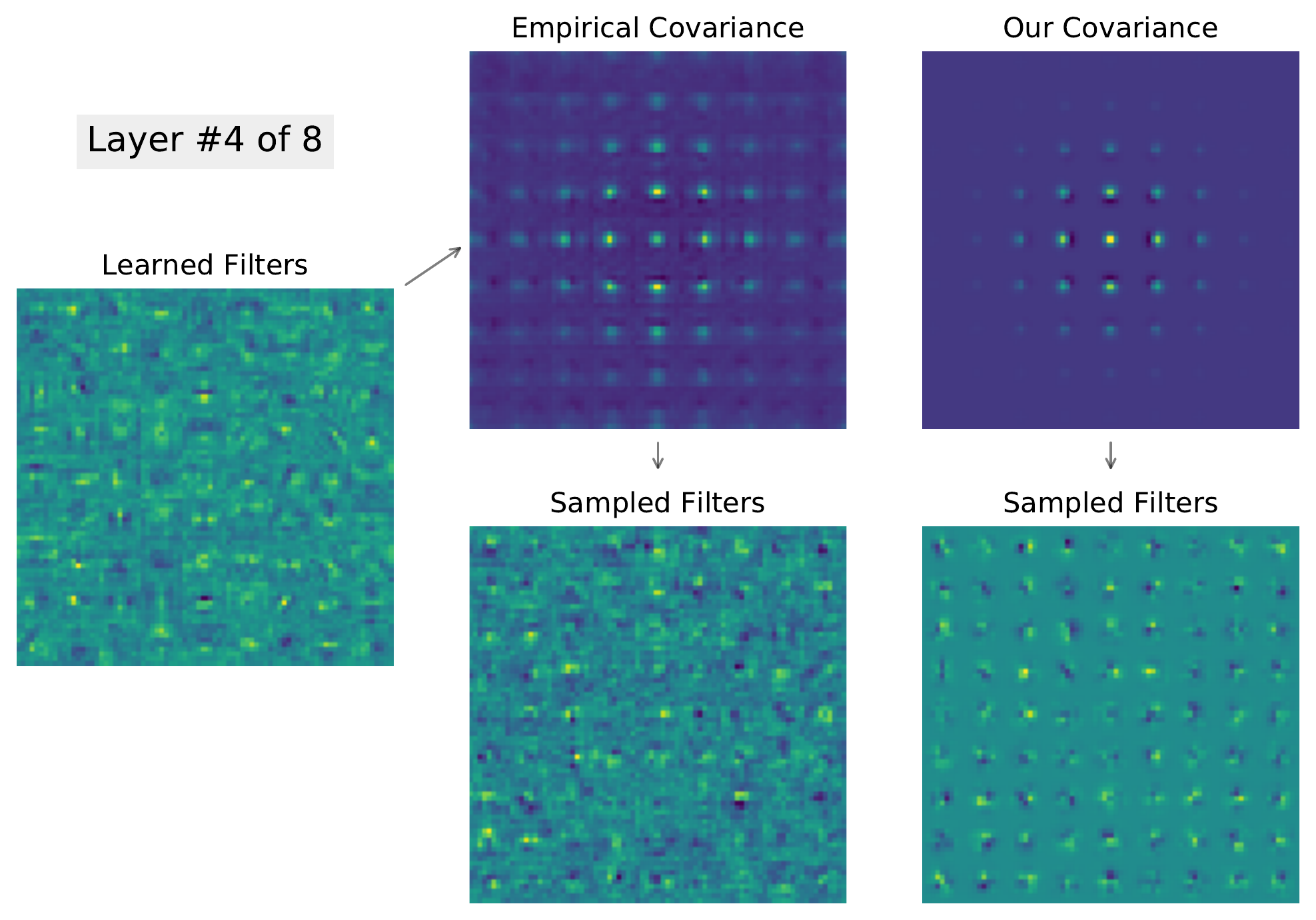}}
    \subfigure{\includegraphics[width=0.7\textwidth]{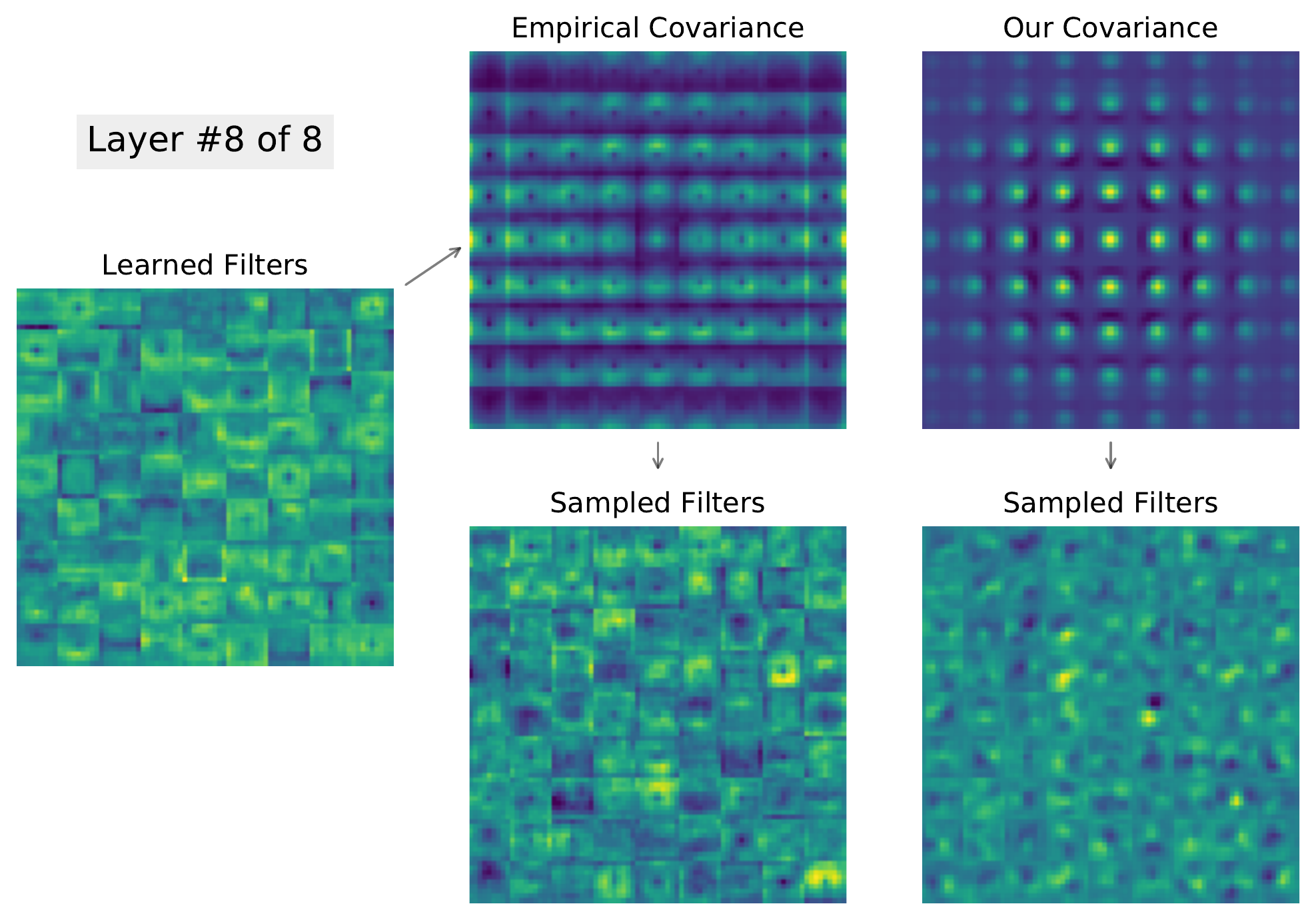}}
    \caption{Filters learned or generated for ConvMixer-256/8 with $2 \times 2$ patches and $9\times 9$ filters trained on CIFAR-10: learned filters (\emph{left}), filters sampled from the Gaussian defined by the empirical covariance matrix of learned filters (\emph{center}),
    and filters from our initialization technique (\emph{right}).}
    \label{fig:example-filters}
\end{figure}

\clearpage

\begin{wrapfigure}[8]{r}{0.25\textwidth}
 \vspace*{-1em}
  \begin{center}
	  \includegraphics[width=0.25\textwidth,trim={50 20 5 0},clip]{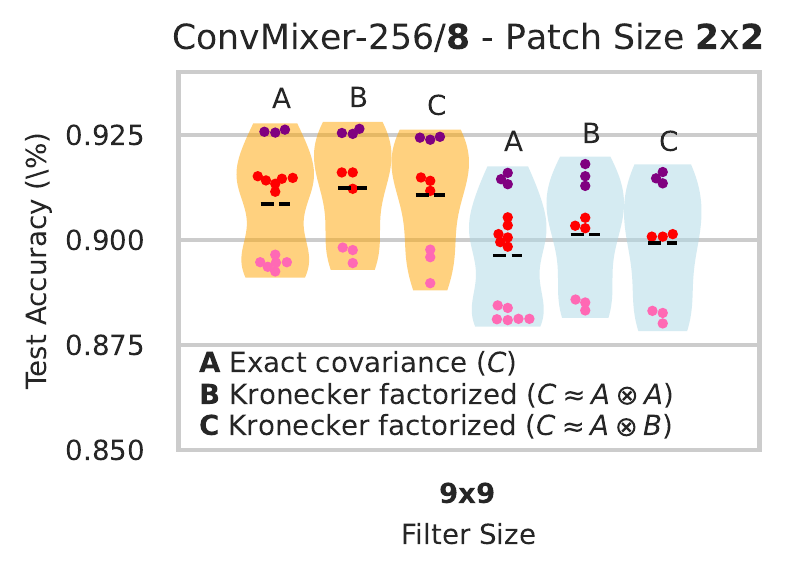}
  \end{center}
	\vspace*{-1.5em}
	\caption{Kronecker-factorized covariances.}
  \label{fig:kronecker}
\end{wrapfigure}

\paragraph{Covariance structure.}
As a first step towards modeling the structure of filter covariances,
we replaced covariances with their Kronecker-factorized counterparts
using the rearranged form of the covariance matrix defined in Eq. (\ref{eq:rearr}),
\emph{i.e.,} $\Sigma = A \otimes A$ where $A \in \mathbb{R}^{k \times k}$.
Surprisingly, this slightly improved performance over unfactorized
covariance transfer (see Fig. ~\ref{fig:kronecker}), suggesting
that filter covariances are not only eminently transferrable
for initialization, but that their core structure may be simpler than meets
the eye.
Kronecker factorizations were computed via gradient descent
minimizing the mean squared error.

\section{Hyperparameter Grid Searches \& Experimental Setup}
\label{sec:hyper}

\begin{figure}[h!]
	\centering
    \includegraphics[width=\textwidth]{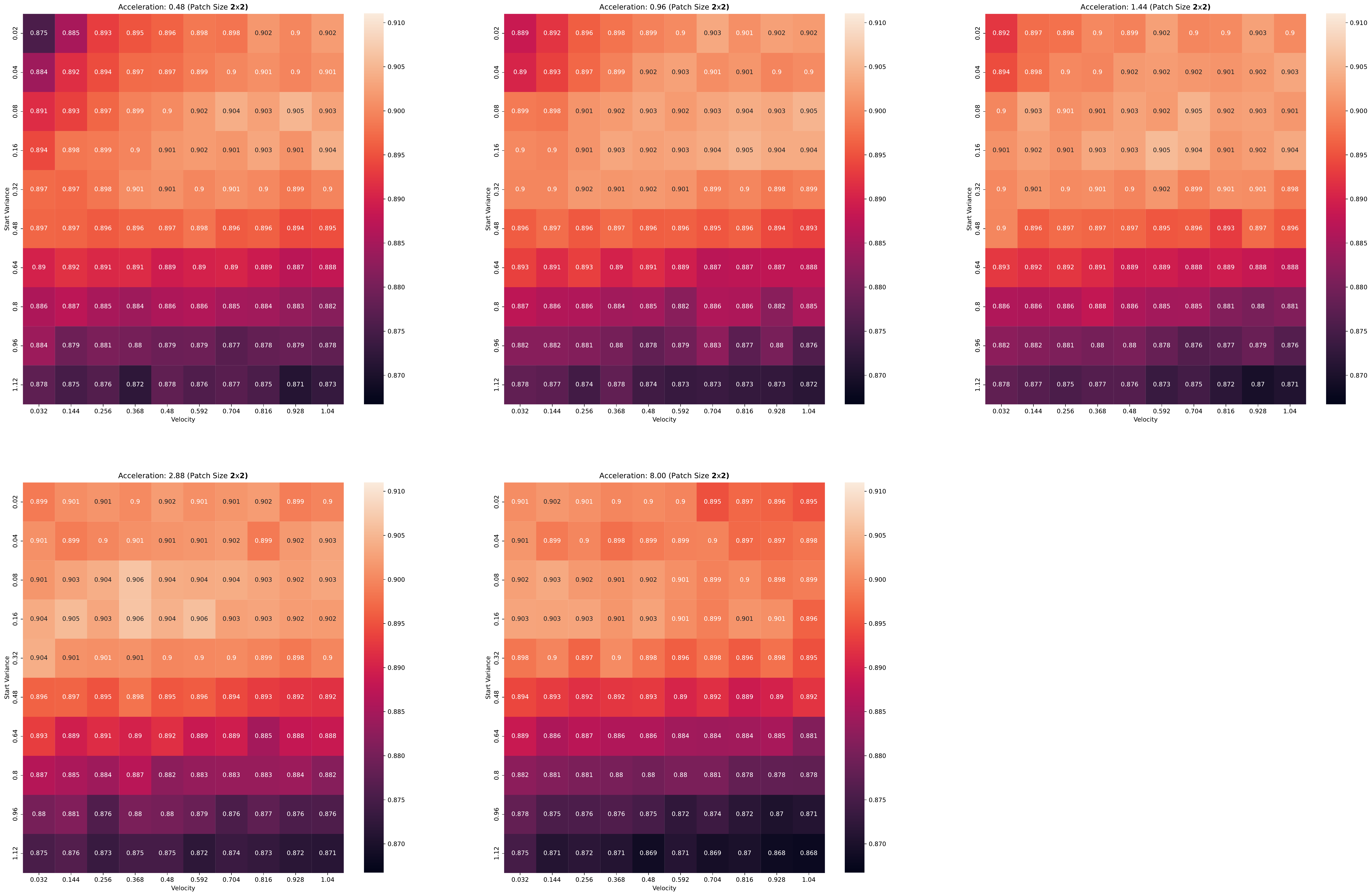}
    \vspace*{-1em}
	\caption{Grid search over initialization parameters $\sigma_0, v_\sigma, a_\sigma$
    for ConvMixer-258/8 with $9\times 9$ \emph{frozen} filters and $2\times 2$ patches
    trained for 20 epochs on CIFAR-10. Note that the performance of uniform initialization
    is only $\approx$85\%, i.e., almost all choices result in \emph{some} improvement.}
	\label{fig:grid-cifar-p1}
\end{figure}

\paragraph{CIFAR-10 hyperparameter search.} 
We chose an initial setting of our method's three hyperparameters
via visual inspection, and then refined them via small-scale grid searches.
For CIFAR-10 experiments, we searched over parameters
for ConvMixer-256/8 with frozen $9 \times 9$ filters trained for 20 epochs,
and chose $\sigma_0=.08, v_\sigma=.37, a_\sigma=2.9$ for $2 \times 2$-patch models,
and found the optimal parameters for $1 \times 1$-patch models to be approximately
doubled. However, note that our initialization is quite robust to different parameter settings,
with the difference from our doubling choice being less than $0.1\%$
(see Figure~\ref{fig:grid-cifar-p1}).
We used the same parameters across all kernel sizes, as well as for ConvNeXt, 
a choice which is likely sub-optimal; our search only serves as a rough heuristic.

\paragraph{ImageNet-1k hyperparameter search.}
\looseness=-1
We did a small grid search using a ConvMixer-512/12
with $14\times 14$ patches and $9\times9$ filters
trained for 10 epochs on ImageNet-1k
(see Appendix~\ref{apx:imnet-exp}),
from which we chose two candidate settings:
$\sigma_0 = .15, v_\sigma = .5, a_\sigma = .25$
for frozen-filter models and $\sigma_0 = .15, v_\sigma = 0.25, a_\sigma = 1.0$
for thawed models.
We use these parameters for all the ImageNet experiments,
even for models with different patch and kernel sizes
(\emph{e.g.,} ConvNeXt).
This demonstrates that \emph{hyperparameter
tuning is optional} for our technique;
its transferability is not surprising
given our results in Sec.~\ref{sec:cov-transfer}.

\section{Implementation}

\begin{figure}[h]
  \begin{minipage}[c]{0.52\textwidth}
{\scriptsize
      \begin{minted}[linenos]{python}
def ConvCov(k, s):
  C = np.zeros((k**2,)*2)
  for i, j in np.ndindex(k,k):
    C[k*i:k*i+k,k*j:k*j+k] = Gauss(k,j,i,s)
  Z,l = Gauss(k,k//2,k//2,s),np.ones((k,k))
  S, M = np.kron(l, Z),np.kron(Z, l)
  return 0.5 * (M * (C - S) + C * S)
\end{minted}
}
  \end{minipage}
  \begin{minipage}[c]{0.4\textwidth}
{\scriptsize
      \begin{minted}[linenos]{python}
def Gauss(k, mx, my, s):
  res = np.zeros((k, k))
  for i, j in np.ndindex(k,k):
    cx,cy = (j-mx-k//2-1)%k,(i-my-k//2-1)%k
    z = ((cx-k//2)**2+(cy-k//2)**2)/s
    res[i, j] = np.exp(-0.5*z)
  return res.reshape(k, k)
\end{minted}
}
  \end{minipage}\hfill
    \caption{Implementation of our convolution covariance construction in NumPy.}
    \label{fig:code}
\end{figure}

\begin{figure}[h]
{\scriptsize
\begin{minted}[linenos]{python}
def Initialize(wconv, d, s0, sv, sa):
  c, _, ks, _ = wconv.shape
  s = s0 + sv * d + 0.5 * sa * d**2
  cov = ConvCov(ks, s).reshape((ks,)*4).transpose(0,2,1,3).reshape((ks**2,)*2)
  filters = np.random.multivariate_normal(np.zeros(ks**2), cov, size=(c,))
  wconv.data = torch.tensor(filters.reshape(c,1,ks,ks),dtype=wconv.dtype,device=wconv.device)
    
# Find depthwise convolutional layers
convs = [x for x in model.modules() if isinstance(x, nn.Conv2d) \
  and len(x.weight.shape) == 4 and x.weight.shape[1] == 1]

# Initialize them according to variance schedule
for i, conv in enumerate(convs):
  Initialize(conv.weight, i / (len(convs) - 1), 0.16, 0.32, 2.88)
\end{minted}
    }
    \caption{Code to use our covariance construction and variance schedule to initalize depthwise convolutional layers in PyTorch.
    \texttt{wconv} is the weight of a depthwise convolutional layer (\texttt{nn.Conv2d}), and $\texttt{d} \in [0, 1]$ is its depth as a fraction of the total depth.}
    \label{fig:all-code}
\end{figure}

\clearpage
\subsection{CIFAR-10 Grid Searches}

\label{apx:cifar-grid}

\begin{figure}[h!]
	\centering
    \includegraphics[width=\textwidth]{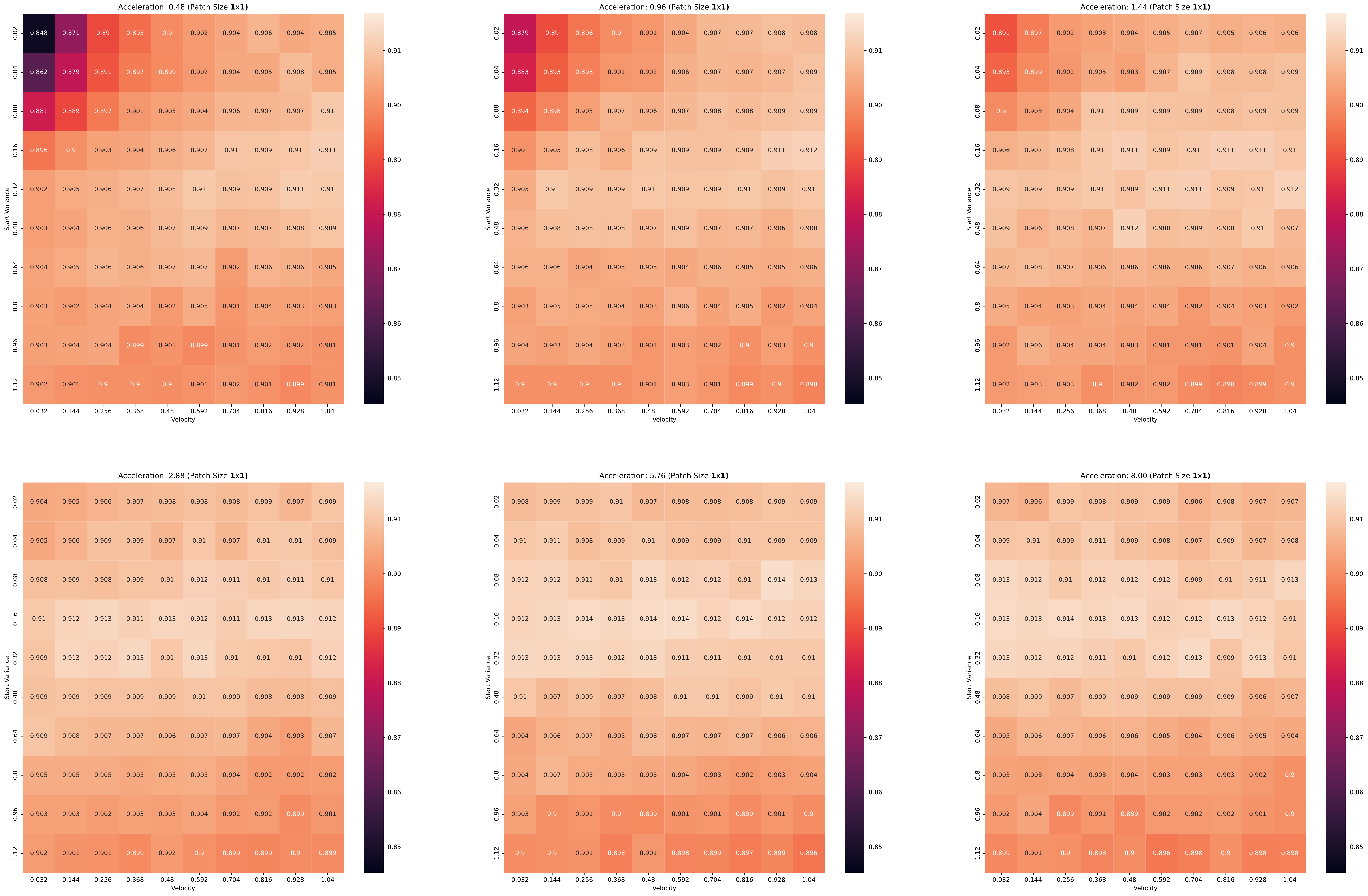}
    \vspace*{-1em}
	\caption{Grid search over initialization parameters $\sigma_0, v_\sigma, a_\sigma$
    for ConvMixer-258/8 with $9\times 9$ \emph{frozen} filters and $1\times 1$ patches
    trained for 20 epochs on CIFAR-10.
    Note that the performance of uniform initialization
    is only $\approx$88\%, i.e., almost all choices result in \emph{some} improvement.}
    
	\label{fig:grid-cifar-p2}
\end{figure}

\begin{figure}[h!]
	\centering
    \includegraphics[width=\textwidth]{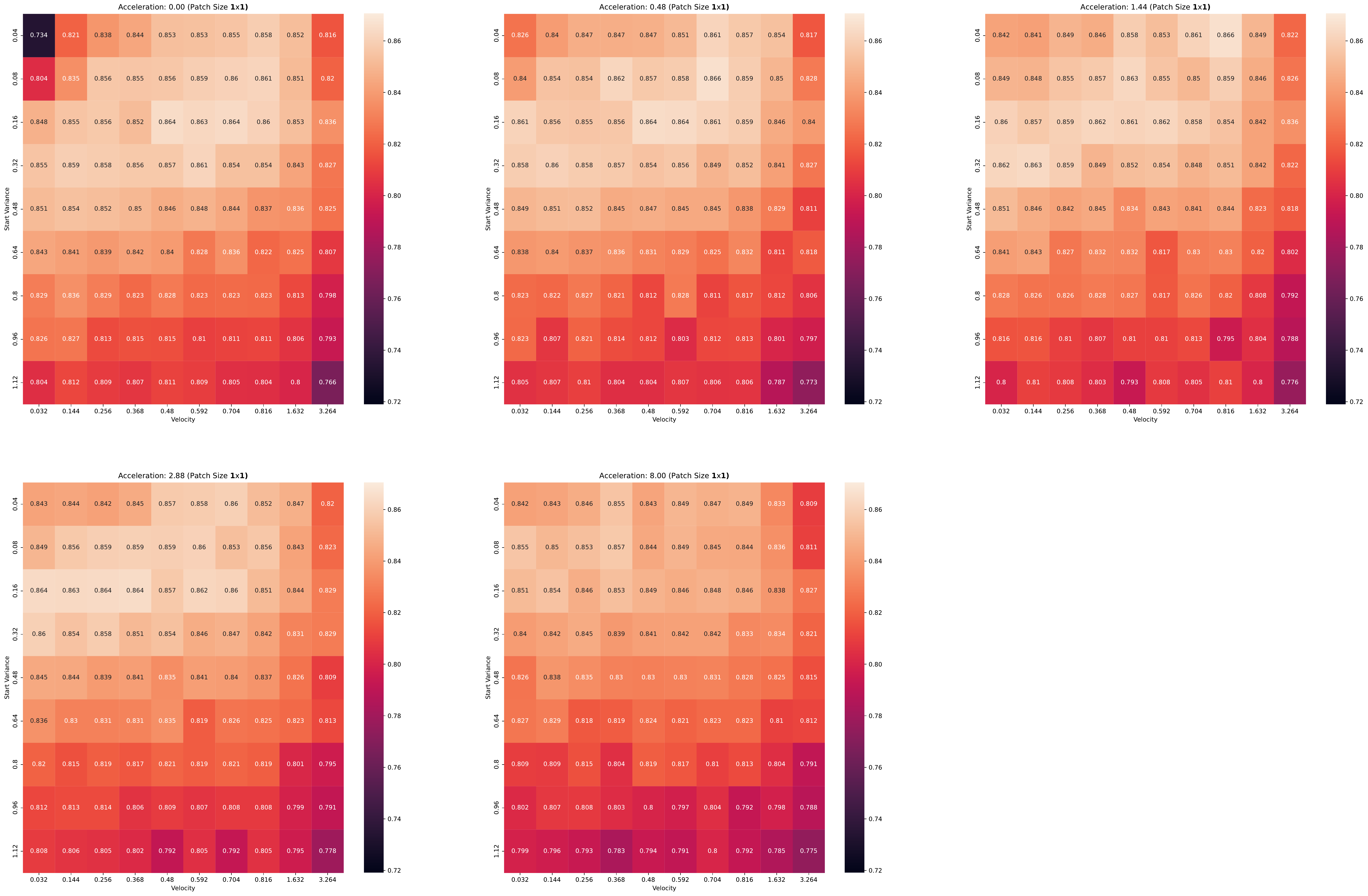}
    \vspace*{-1em}
	\caption{Grid search over initialiation parameters $\sigma_0, v_\sigma, a_\sigma$
    for ConvNeXt-atto on CIFAR-10 with frozen filters and $1\times 1$ patches trained for 20 epochs  on CIFAR-10.
    Note the baseline performance with uniform initialization is around $80\%$, \emph{i.e.,}
    compared to ConvMixer there are more potentially disadvantageous parameter combinations.}
	\label{fig:convnext-p1}
\end{figure}

\begin{figure}[h!]
	\centering
    \includegraphics[width=\textwidth]{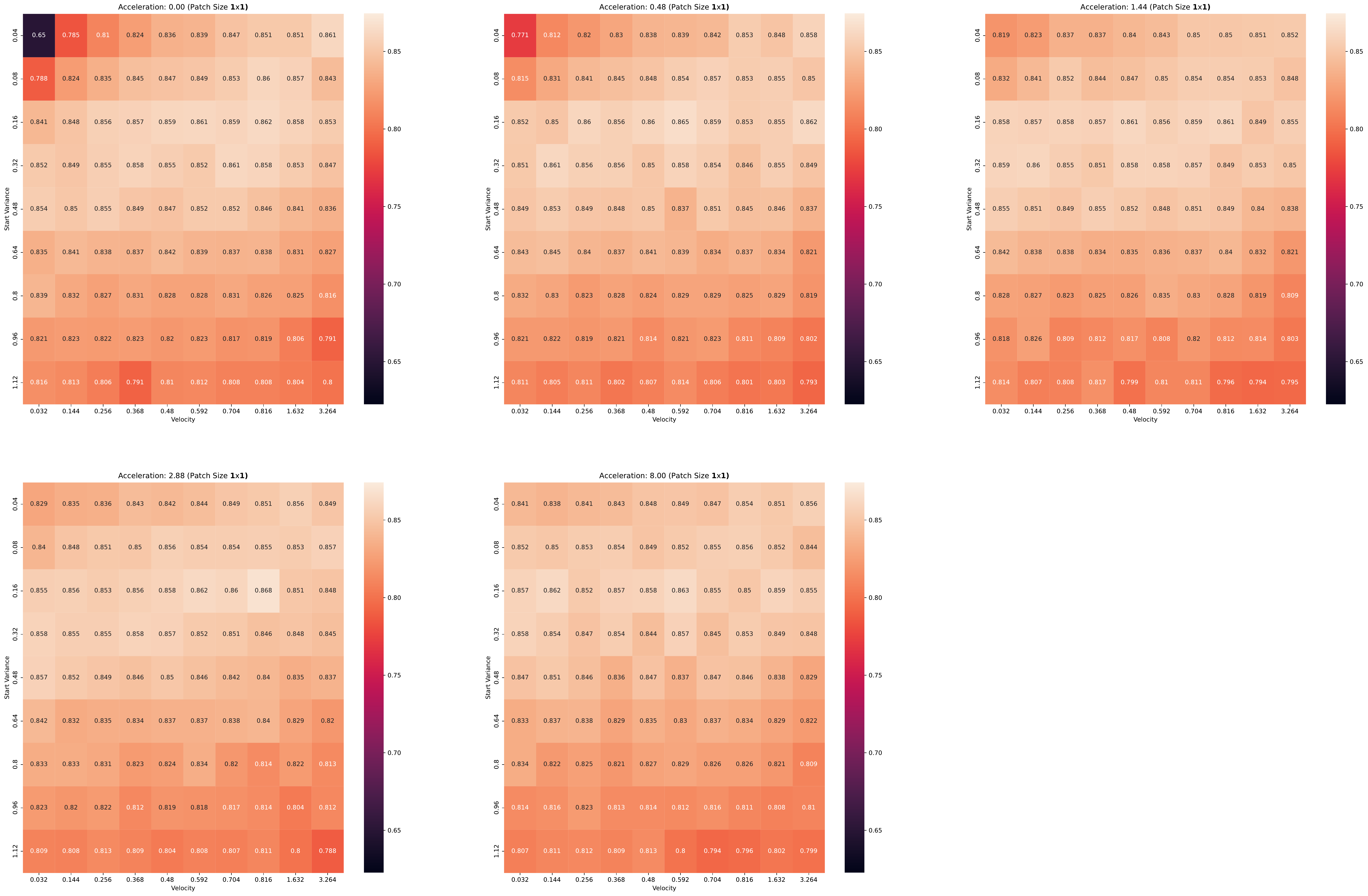}
    \vspace*{-1em}
	\caption{Grid search over initialiation parameters $\sigma_0, v_\sigma, a_\sigma$
    for ConvNeXt-atto on CIFAR-10 with frozen filters and $1\times 1$ patches trained for 20 epochs,
    using the ``sawtooth'' variance schedule (see Fig~\ref{fig:sawtooth}) to account for downsampling layers.
    While this perhaps shows better robustness to parameter changes than Fig.~\ref{fig:convnext-p1},
    the effect could also be due to effectively dividing the parameters by two.}
	\label{fig:convnext-p1-saw}
\end{figure}

\begin{figure}[h!]
	\centering
    \includegraphics[width=\textwidth]{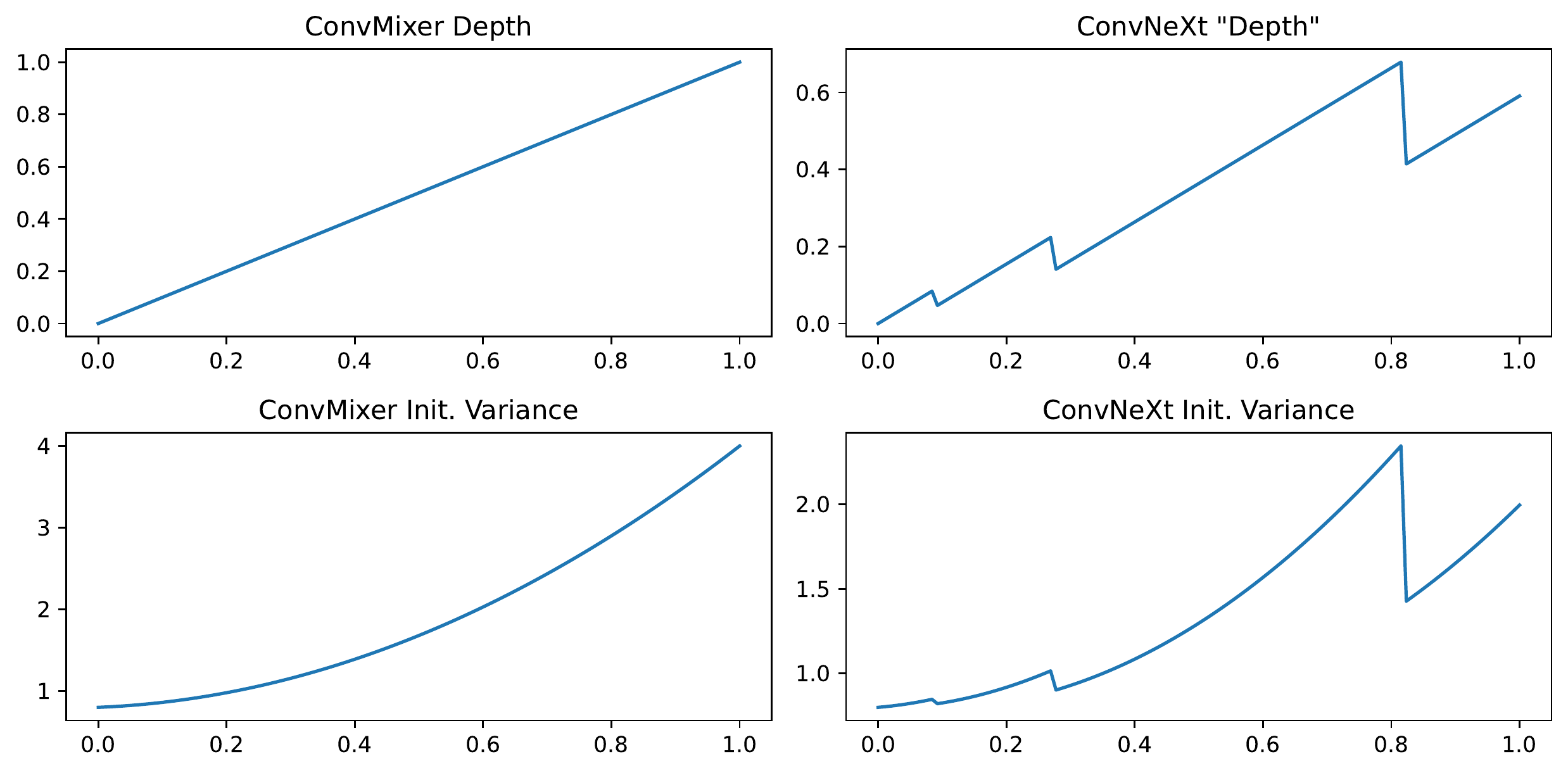}
    \vspace*{-1em}
    \caption{Proposed stepwise variance schedule for ConvNeXt, \emph{i.e.,} a model including downsampling
    layers. In our experiments, we saw no advantage to using this scheme.}
	\label{fig:sawtooth}
\end{figure}

\clearpage

\subsection{ImageNet Grid Searches}

\begin{figure}[h!]
	\centering
    \includegraphics[width=\textwidth]{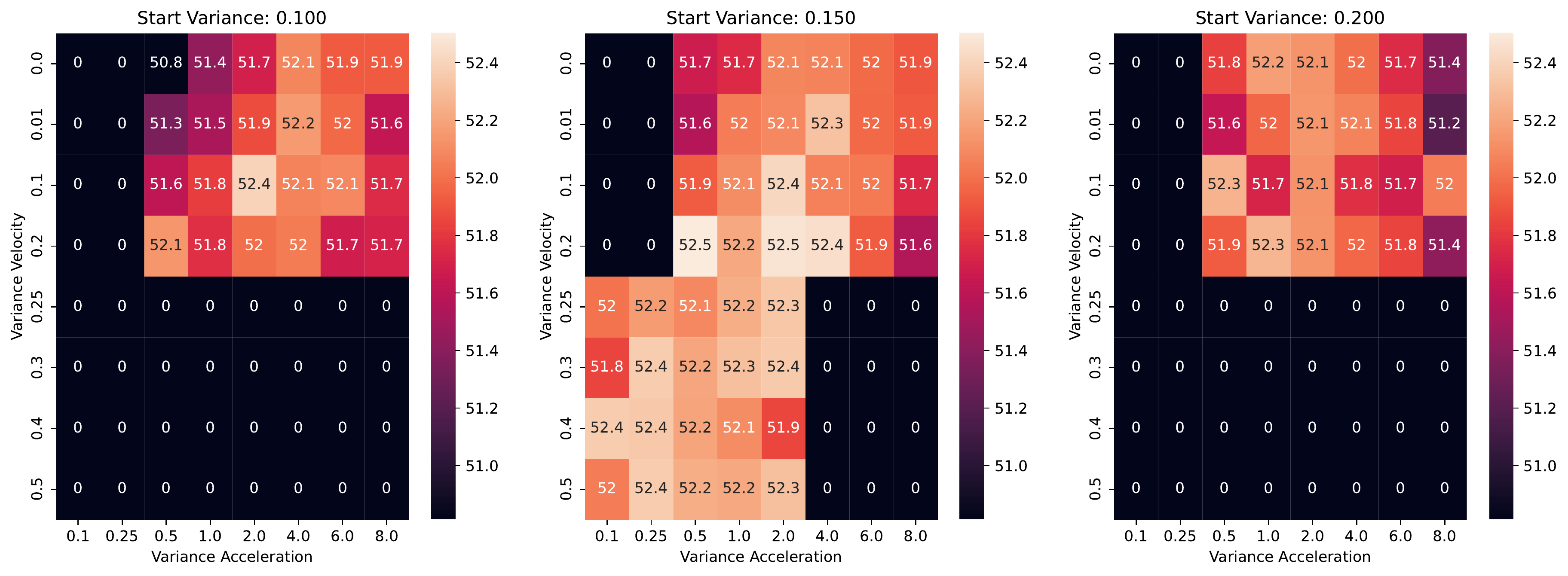}
    \vspace*{-1em}
    \caption{\textbf{Frozen filters:} Grid search over initialization parameters for ConvMixer-512/12 with $14\times 14$ patches
    and $9 \times 9$ filters, 10 epochs. Zeros indicate that the experiment did not run.}
	\label{fig:grid-imnet}
\end{figure}

\begin{figure}[h!]
	\centering
    \includegraphics[width=\textwidth]{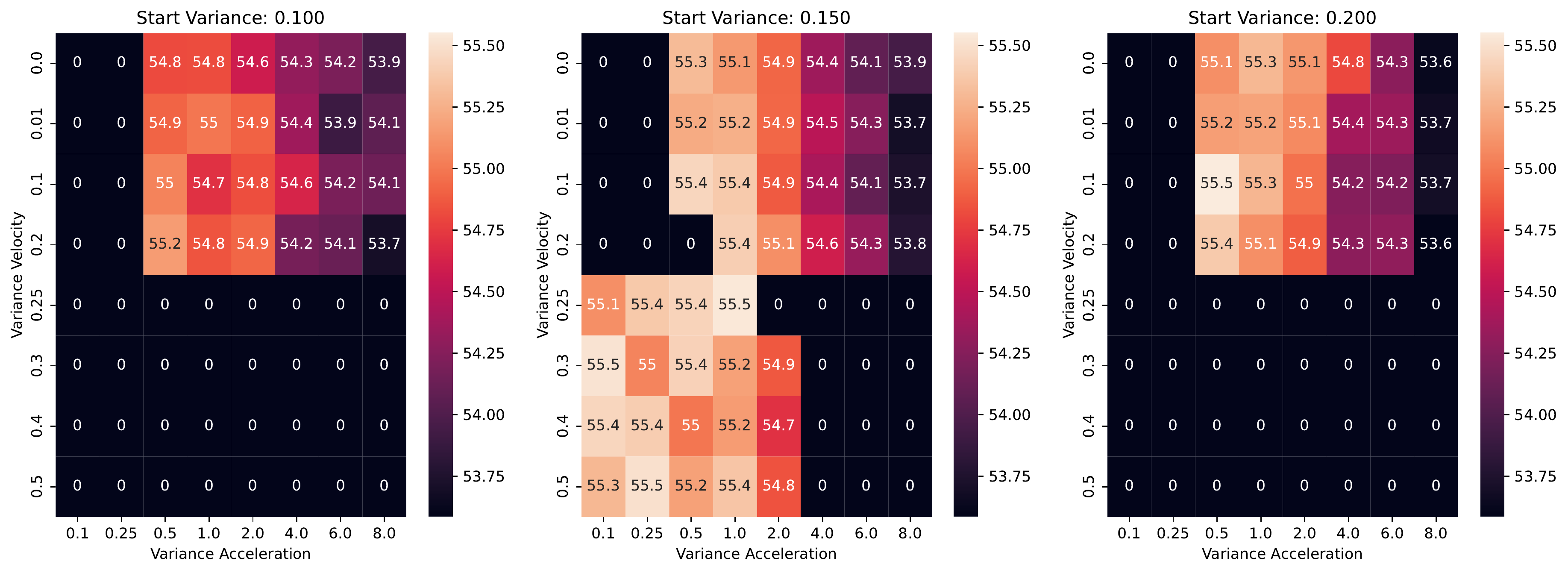}
    \vspace*{-1em}
    \caption{\textbf{Thawed filters:} Grid search over initialization parameters for ConvMixer-512/12 with $14\times 14$ patches
    and $9 \times 9$ filters, 10 epochs.}
	\label{fig:grid-imnet}
\end{figure}

\clearpage

\section{Shift Function Definition \& Proof}
\label{apx:shift-proof}

For a given matrix $Z \in \mathbb{R}^{k \times k}$ (\emph{e.g.,} a Gaussian kernel centered at the top left of the filter),
we define the Shift operator as follows:
\begin{equation}
    \text{Shift}(Z, \Delta x, \Delta y)_{i,j} = Z_{(i + \Delta x) \text{ mod } k, (j + \Delta y) \text{ mod } k}.
\end{equation}
Note that this can be achieved using \texttt{np.roll} in NumPy.
Then, if
\begin{equation}
    [C_{i,j}] = \text{Shift}(Z_{\sigma}, i, j)
\end{equation}
and the operation $(.)^B$ is defined by
\begin{equation}
\Sigma^B = \Sigma' \Longleftrightarrow \left[ \Sigma_{i,j}\right]_{\ell, m} = \left[ \Sigma^\prime_{\ell, m}\right]_{i,j}  \text{ for } 1 \leq i, j, \ell, m \leq k,
\end{equation}
then
\begin{align}
    [C_{i,j}]_{\ell, m} &= \text{Shift}(Z, i, j)_{\ell,m} = Z_{(i + \ell) \text{ mod } k, (j + m) \text{ mod } k}\\
    &= Z_{(\ell + i) \text{ mod } k, (m + j) \text{ mod } k} = \text{Shift}(Z, \ell, m)_{i, j} = [C_{\ell,m}]_{i, j},
\end{align}
which shows that $\left[ C_{i,j}\right]_{\ell, m} = \left[ C_{\ell, m}\right]_{i,j}$ for all $1 \leq i, j, \ell, m \leq k$, \emph{i.e.},
$C$ is ``block-symmetric'', or $C = C^B$. $\square$

\clearpage

\section{Additional ImageNet Experiments}
\label{apx:imnet-exp}

\begin{figure}[h]
  \begin{minipage}[c]{0.33\textwidth}
    \caption{
        Covariance matrices
        from a ConvMixer trained on ImageNet
        exhibit similar structure to
        those of ConvMixers trained on CIFAR-10;
        however, later layers tend to have more structure,
        including a ``checkerboard'' pattern in each sub-block.
    } \label{fig:imnet-covs}
  \end{minipage}
  \begin{minipage}[c]{0.67\textwidth}
    \includegraphics[width=\textwidth]{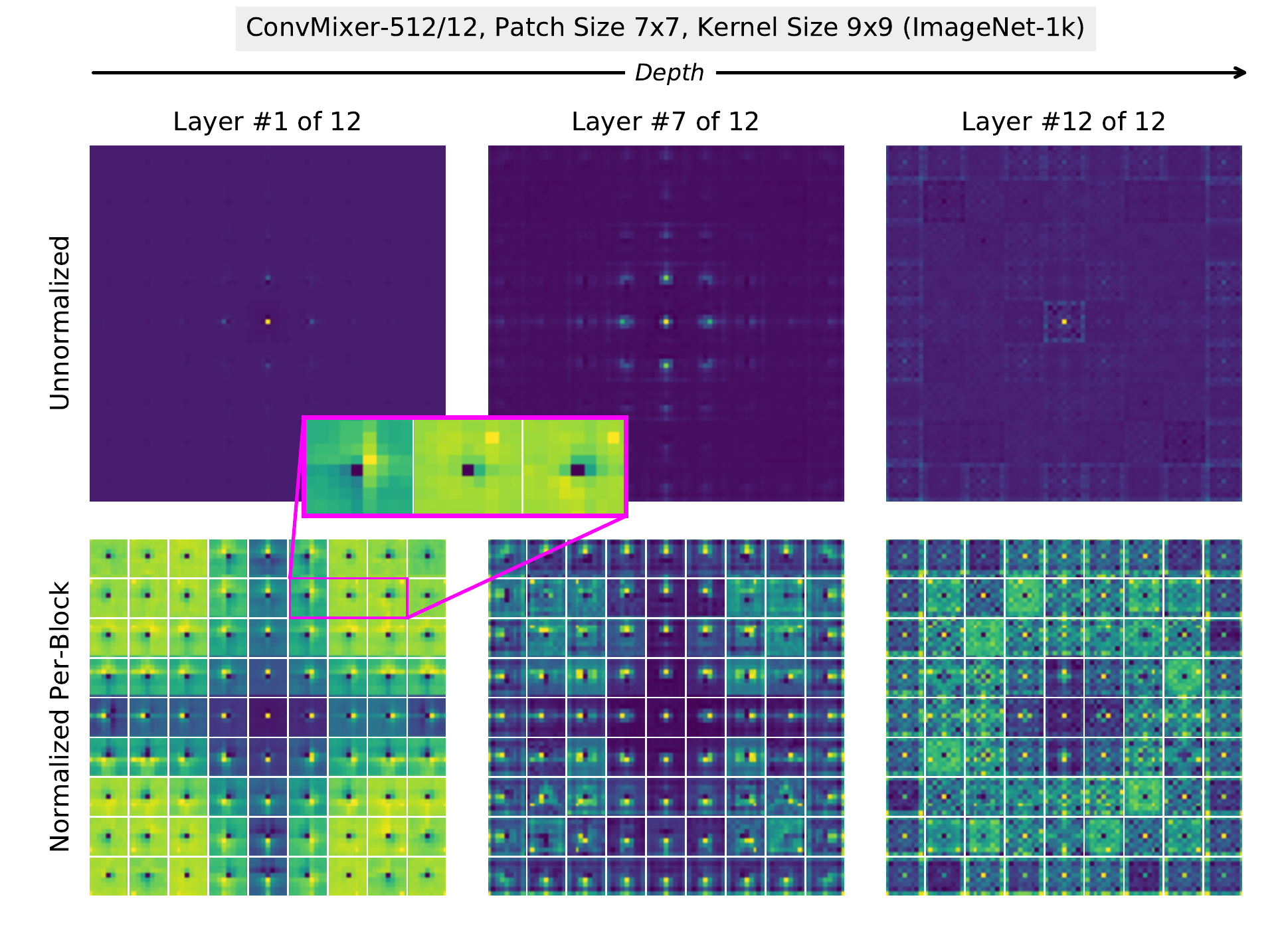}
  \end{minipage}\hfill
\end{figure}

\clearpage

\subsection{10-epoch ImageNet Results}

\begin{table}[h]
    \centering
\begin{tabular}{l|l|l}
\textbf{ConvMixer-512/12: Patch Size 14, Kernel Size 9} & Thawed & Frozen  \\
        \hline
Uniform init & 54.5  & 47.4   \\
Stats from CM-512/12      & 55.5  & 53.4     \\
Stats from CM-64/12      & 55.2  & 52.7   \\
Filters transferred from CM-512/12       & 55.1  & 54.4    \\
\hline
Our init (.15, .3, .5)          & 55.4  & 52.2      \\
Our init (.15, .5, .25)         & 55.5  & 52.4     \\

\end{tabular}
    \caption{ConvMixer performance on ImageNet-1k training with 10 epochs. Our initialization
    performs comparably to loading covariance matrices from previously-trained models (which were trained for 150 epochs).}

\end{table}

\begin{table}[h!]
    \centering
\begin{tabular}{l|l|l}
\textbf{ConvMixer-512/12: Patch Size 7, Kernel Size 9} & Thawed & Frozen  \\
        \hline
Uniform init                             & 61.87  & 56.73   \\
Stats from CM-512/12                     & 62.56  & 60.79     \\
Stats from CM-64/12                      & 62.72  & 60.86   \\
Filters transferred from CM-512/12       & 62.81  & 61.83    \\
\hline
Our init (.15, .3, .5)          & 62.49  & 58.94      \\
Our init (.15, .5, .25)         & 62.59  & 59.31      \\
\end{tabular}
    \caption{ImageNet 10-epoch training}
\end{table}

\begin{table}[h!]
    \centering
\begin{tabular}{l|l|l}
\textbf{ConvMixer-512/24: Patch Size 14, Kernel Size 9} & Thawed & Frozen  \\
        \hline
Uniform init                             & 50.40  & 43.00   \\
Stats from CM-512/12                     & 53.03  & 51.45     \\
Stats from CM-64/12                      & 53.16  & 51.25   \\
Filters transferred from CM-512/12       & 52.87  & 52.12    \\
\hline
Our init (.15, .3, .5)          & 53.80  & 51.16      \\
Our init (.15, .5, .25)         & 53.76  & 50.81      \\
\end{tabular}
    \caption{ImageNet 10-epoch training}
\end{table}

\begin{table}[h!]
    \centering
\begin{tabular}{l|l|l}
\textbf{ConvNeXt-Atto} & Thawed & Frozen  \\
        \hline
Uniform init                                      & 31.37  & 23.63   \\
Stats from the same arch                          & 33.44  & 40.41     \\
Stats from $1/8^{\text{th}}$-width arch           & 29.81  & 31.47   \\
Filters transferred from same arch                & 31.68  & 40.48    \\
\hline
Our init (.15, .3, .5)                   & 37.64  & 34.59      \\
Our init (.15, .5, .25)                  & 31.34  & 34.23      \\
Our init (.15, .25, 1.0)                 & 38.01  & 33.98      \\
\end{tabular}
    \caption{ImageNet 10-epoch training}
\end{table}

\begin{table}[h!]
    \centering
\begin{tabular}{l|l|l}
\textbf{ConvNeXt-Tiny} & Thawed & Frozen  \\
        \hline
Uniform init                                      & 32.51  & 25.94   \\
Stats from the same arch                          & 42.78  & 41.54     \\
Stats from $1/8^{\text{th}}$-width arch           & 44.60  & 42.86   \\
Filters transferred from same arch                & 31.01  & 45.32    \\
\hline
Our init (.15, .3, .5)                   & 35.64  & 35.04      \\
Our init (.15, .5, .25)                  & 40.17  & 38.91      \\
Our init (.15, .25, 1.0)                 & 40.78  & 36.62      \\
\end{tabular}
    \caption{ImageNet 10-epoch training}
\end{table}

\clearpage

\subsection{50-epoch ImageNet Results}

\begin{table}[h!]
    \centering
\begin{tabular}{l|l|l}
\textbf{ConvNeXt-Atto} & Thawed & Frozen  \\
        \hline
Uniform init                                      & 69.96  & 51.43   \\
Stats from the same arch                          & 68.83  & 66.71     \\
Stats from $1/8^{\text{th}}$-width arch           & 68.69  & 66.31   \\
Filters transferred from same arch                & 68.01  & 67.29    \\
\hline
Our init (.15, .3, .5)                   & 65.55  & 63.48      \\
Our init (.15, .5, .25)                  & 67.84  & 64.52      \\
Our init (.15, .25, 1.0)                 & 68.06  & 63.43      \\
\end{tabular}
    \caption{ImageNet \textbf{50-epoch} training}
\end{table}

\begin{table}[h!]
    \centering
\begin{tabular}{l|l|l}
\textbf{ConvMixer-512/12: Patch Size 14, Kernel Size 9} & Thawed & Frozen  \\
        \hline
Uniform init                                      & 67.03  & 60.47   \\
Stats from the same arch                          & 67.13  & 65.08     \\
Stats from $1/8^{\text{th}}$-width arch           & 66.75  & 64.94   \\
Filters transferred from same arch                & 67.28  & 66.11    \\
\hline
Our init (.15, .3, .5)                   & 66.12  & 64.39      \\
Our init (.15, .5, .25)                  & 67.41  & 64.43      \\
Our init (.15, .25, 1.0)                 & 67.34  & 64.12      \\
\end{tabular}
    \caption{ImageNet \textbf{50-epoch} training}
\end{table}

\begin{table}[h!]
    \centering
\begin{tabular}{l|l|l}
\textbf{ConvMixer-512/24: Patch Size 14, Kernel Size 9} & Thawed & Frozen  \\
        \hline
Uniform init                                      & 67.76  & 62.50   \\
Stats from the same arch                          & 68.92  & 67.91     \\
Stats from $1/8^{\text{th}}$-width arch           & 68.78  & 67.36   \\
Filters transferred from same arch                & 69.42  & 68.66    \\
\hline
Our init (.15, .3, .5)                   & 69.05  & 66.20      \\
Our init (.15, .5, .25)                  & 69.60  & 66.57      \\
Our init (.15, .25, 1.0)                 & 69.52  & 66.38      \\
\end{tabular}
    \caption{ImageNet \textbf{50-epoch} training }
\end{table}

\clearpage

\section{Additional CIFAR-10 Tables}

  \begin{sidewaystable}
    \caption{CIFAR-10 results for ConvMixer-256/8 with patch size 2.  \textbf{Bold} denotes the highest per group,
        and {\color{darkblue} \textbf{blue bold}} denotes the second highest. }
    \label{fig:apx-cm-256-8-2}
    \vspace{0.5em}
        \centering
    {\scriptsize
    \begin{tabular}{|c|c|b|b|b|b|b|a|a|a|a|a|}
        \hline
         &  & \multicolumn{5}{b|}{\textsc{Thawed}} & \multicolumn{5}{a|}{\textsc{Frozen}} \\
        \hline
        \makecell{Filt. \\ Size} & \makecell{\# \\ Eps} & Uniform & Cov. transfer & \makecell{Cov. transfer \\ ($\tfrac{1}{8}$ width)} & Direct transfer & Our init & Uniform & Cov. transfer & \makecell{Cov. transfer \\ ($\tfrac{1}{8}$ width)} & Direct transfer & Our init \\
        \hline
        \multirow{3}{*}{3} & 20 & $89.69 \pm .20$ & $\mathbf{90.08 \pm .39}$ & $89.76 \pm .14$ & $\color{darkblue}  \mathbf{90.02 \pm .14}$ & $89.64 \pm .13$ & $88.99 \pm .41$ & $\color{darkblue} \mathbf{89.56 \pm .11}$ & $89.54 \pm .40$ & $\mathbf{89.94 \pm .08}$ & $88.94 \pm .25$ \\
& 50 & $\color{darkblue}  \mathbf{91.92 \pm .03}$ & $\mathbf{92.01 \pm .21}$ & $91.83 \pm .02$ & $91.69 \pm .18$ & $91.63 \pm .07$ & $91.12 \pm .22$ & $\color{darkblue} \mathbf{91.42 \pm .15}$ & $91.34 \pm .27$ & $\mathbf{91.62 \pm .17}$ & $91.02 \pm .25$ \\
& 200 & $\mathbf{93.08 \pm .16}$ & $92.94 \pm .26$ & $92.92 \pm .19$ & $\color{darkblue}  \mathbf{92.97 \pm .18}$ & $92.84 \pm .11$ & $92.38 \pm .27$ & $\mathbf{92.55 \pm .09}$ & $92.27 \pm .21$ & $\color{darkblue} \mathbf{92.47 \pm .19}$ & $92.09 \pm .13$ \\
\hline
\multirow{3}{*}{7} & 20 & $89.66 \pm .21$ & $89.91 \pm .09$ & $\color{darkblue}  \mathbf{90.63 \pm .10}$ & $90.48 \pm .26$ & $\mathbf{90.79 \pm .24}$ & $86.73 \pm .27$ & $88.86 \pm .36$ & $89.36 \pm .09$ & $\color{darkblue} \mathbf{90.00 \pm .27}$ & $\mathbf{90.22 \pm .03}$ \\
& 50 & $91.81 \pm .15$ & $91.77 \pm .26$ & $91.88 \pm .19$ & $\color{darkblue}  \mathbf{91.91 \pm .20}$ & $\mathbf{92.48 \pm .08}$ & $89.44 \pm .23$ & $90.72 \pm .13$ & $91.05 \pm .12$ & $\color{darkblue} \mathbf{91.54 \pm .37}$ & $\mathbf{92.02 \pm .23}$ \\
& 200 & $\color{darkblue}  \mathbf{92.86 \pm .15}$ & $92.77 \pm .28$ & $92.74 \pm .15$ & $92.72 \pm .06$ & $\mathbf{93.40 \pm .20}$ & $90.70 \pm .12$ & $91.68 \pm .16$ & $91.84 \pm .05$ & $\color{darkblue} \mathbf{92.36 \pm .20}$ & $\mathbf{92.83 \pm .24}$ \\
\hline
\multirow{3}{*}{9} & 20 & $89.26 \pm .35$ & $89.70 \pm .06$ & $\color{darkblue}  \mathbf{90.22 \pm .36}$ & $90.14 \pm .06$ & $\mathbf{90.85 \pm .14}$ & $84.97 \pm .03$ & $88.18 \pm .12$ & $\color{darkblue} \mathbf{89.67 \pm .07}$ & $89.50 \pm .37$ & $\mathbf{90.56 \pm .09}$ \\
& 50 & $91.74 \pm .12$ & $91.63 \pm .15$ & $\color{darkblue}  \mathbf{92.11 \pm .18}$ & $91.79 \pm .18$ & $\mathbf{92.54 \pm .16}$ & $88.25 \pm .39$ & $90.22 \pm .09$ & $91.01 \pm .12$ & $\color{darkblue} \mathbf{91.23 \pm .13}$ & $\mathbf{91.96 \pm .12}$ \\
& 200 & $92.65 \pm .16$ & $92.53 \pm .13$ & $\color{darkblue}  \mathbf{92.85 \pm .12}$ & $92.62 \pm .26$ & $\mathbf{93.21 \pm .09}$ & $90.04 \pm .33$ & $91.34 \pm .08$ & $92.00 \pm .18$ & $\color{darkblue} \mathbf{92.05 \pm .28}$ & $\mathbf{93.00 \pm .05}$ \\
\hline
\multirow{3}{*}{15} & 20 & $86.64 \pm .30$ & $88.19 \pm .51$ & $\color{darkblue}  \mathbf{89.27 \pm .05}$ & $89.17 \pm .19$ & $\mathbf{90.33 \pm .22}$ & $81.99 \pm .21$ & $86.31 \pm .15$ & $87.52 \pm .23$ & $\color{darkblue} \mathbf{88.39 \pm .28}$ & $\mathbf{90.38 \pm .06}$ \\
& 50 & $89.94 \pm .45$ & $90.26 \pm .16$ & $\color{darkblue}  \mathbf{90.81 \pm .26}$ & $90.79 \pm .24$ & $\mathbf{92.17 \pm .23}$ & $85.05 \pm .33$ & $88.71 \pm .08$ & $89.51 \pm .06$ & $\color{darkblue} \mathbf{90.08 \pm .09}$ & $\mathbf{92.11 \pm .24}$ \\
& 200 & $91.79 \pm .23$ & $91.60 \pm .21$ & $\color{darkblue}  \mathbf{92.01 \pm .19}$ & $91.87 \pm .26$ & $\mathbf{92.94 \pm .11}$ & $87.64 \pm .05$ & $89.93 \pm .25$ & $90.61 \pm .17$ & $\color{darkblue} \mathbf{90.94 \pm .12}$ & $\mathbf{93.02 \pm .09}$ \\
\hline

    \end{tabular}
        }

    \end{sidewaystable}

    \begin{sidewaystable}
    \caption{CIFAR-10 results for ConvMixer-256/24 with patch size 2. \textbf{Bold} denotes the highest per group,
        and {\color{darkblue} \textbf{blue bold}} denotes the second highest.  }
    \label{fig:apx-cm-256-24-2}
    \vspace{0.5em}
        \centering
    {\scriptsize
    \begin{tabular}{|c|c|b|b|b|b|b|a|a|a|a|a|}
        \hline
         &  & \multicolumn{5}{b|}{\textsc{Thawed}} & \multicolumn{5}{a|}{\textsc{Frozen}} \\
        \hline
        \makecell{Filt. \\ Size} & \makecell{\# \\ Eps} & Uniform & Cov. transfer & \makecell{Cov. transfer \\ ($\tfrac{1}{8}$ width)} & Direct transfer & Our init & Uniform & Cov. transfer & \makecell{Cov. transfer \\ ($\tfrac{1}{8}$ width)} & Direct transfer & Our init \\
        \hline
        \multirow{3}{*}{3} & 20 & $88.37 \pm .11$ & $88.79 \pm .07$ & $\color{darkblue}  \mathbf{88.97 \pm .13}$ & $\mathbf{89.22 \pm .26}$ & $88.42 \pm .14$ & $87.74 \pm .32$ & $88.63 \pm .20$ & $\color{darkblue} \mathbf{88.80 \pm .12}$ & $\mathbf{88.95 \pm .11}$ & $88.05 \pm .09$ \\
& 50 & $92.38 \pm .15$ & $92.30 \pm .24$ & $\color{darkblue}  \mathbf{92.46 \pm .07}$ & $\mathbf{92.56 \pm .19}$ & $92.19 \pm .24$ & $91.86 \pm .18$ & $92.02 \pm .32$ & $\mathbf{92.37 \pm .03}$ & $\color{darkblue} \mathbf{92.31 \pm .14}$ & $92.00 \pm .23$ \\
& 200 & $94.13 \pm .07$ & $94.32 \pm .12$ & $\mathbf{94.41 \pm .11}$ & $\color{darkblue}  \mathbf{94.37 \pm .03}$ & $94.16 \pm .20$ & $93.71 \pm .15$ & $93.91 \pm .13$ & $\mathbf{94.28 \pm .20}$ & $\color{darkblue} \mathbf{93.95 \pm .23}$ & $93.84 \pm .10$ \\
\hline
\multirow{3}{*}{7} & 20 & $88.49 \pm .46$ & $89.08 \pm .15$ & $\color{darkblue}  \mathbf{89.90 \pm .14}$ & $89.81 \pm .16$ & $\mathbf{90.28 \pm .18}$ & $85.81 \pm .05$ & $87.98 \pm .08$ & $89.19 \pm .12$ & $\color{darkblue} \mathbf{89.34 \pm .40}$ & $\mathbf{90.09 \pm .22}$ \\
& 50 & $91.90 \pm .17$ & $91.73 \pm .26$ & $\color{darkblue}  \mathbf{92.39 \pm .17}$ & $92.25 \pm .12$ & $\mathbf{93.15 \pm .08}$ & $89.94 \pm .17$ & $90.86 \pm .07$ & $91.56 \pm .09$ & $\color{darkblue} \mathbf{91.91 \pm .08}$ & $\mathbf{92.80 \pm .27}$ \\
& 200 & $93.57 \pm .08$ & $93.43 \pm .21$ & $\color{darkblue}  \mathbf{93.71 \pm .20}$ & $93.62 \pm .19$ & $\mathbf{94.44 \pm .26}$ & $91.78 \pm .22$ & $92.78 \pm .16$ & $\color{darkblue} \mathbf{93.24 \pm .25}$ & $93.00 \pm .25$ & $\mathbf{94.03 \pm .23}$ \\
\hline
\multirow{3}{*}{9} & 20 & $87.99 \pm .41$ & $88.25 \pm .13$ & $\color{darkblue}  \mathbf{89.44 \pm .03}$ & $89.42 \pm .26$ & $\mathbf{90.54 \pm .15}$ & $83.42 \pm .33$ & $87.38 \pm .18$ & $88.46 \pm .45$ & $\color{darkblue} \mathbf{88.90 \pm .52}$ & $\mathbf{90.31 \pm .12}$ \\
& 50 & $91.06 \pm .11$ & $91.36 \pm .17$ & $\color{darkblue}  \mathbf{91.87 \pm .11}$ & $91.69 \pm .06$ & $\mathbf{93.03 \pm .10}$ & $88.38 \pm .08$ & $90.25 \pm .34$ & $\color{darkblue} \mathbf{91.12 \pm .16}$ & $91.04 \pm .18$ & $\mathbf{92.78 \pm .27}$ \\
& 200 & $93.12 \pm .37$ & $93.08 \pm .17$ & $\color{darkblue}  \mathbf{93.42 \pm .21}$ & $93.16 \pm .21$ & $\mathbf{94.12 \pm .18}$ & $90.86 \pm .07$ & $91.86 \pm .28$ & $\color{darkblue} \mathbf{92.60 \pm .09}$ & $92.53 \pm .15$ & $\mathbf{94.03 \pm .09}$ \\
\hline
\multirow{3}{*}{15} & 20 & $84.95 \pm .50$ & $85.80 \pm .48$ & $86.67 \pm .29$ & $\color{darkblue}  \mathbf{87.69 \pm .57}$ & $\mathbf{90.08 \pm .23}$ & $80.72 \pm .20$ & $84.27 \pm .41$ & $85.70 \pm .25$ & $\color{darkblue} \mathbf{86.75 \pm .52}$ & $\mathbf{90.03 \pm .06}$ \\
& 50 & $89.74 \pm .11$ & $89.68 \pm .15$ & $90.10 \pm .12$ & $\color{darkblue}  \mathbf{90.22 \pm .27}$ & $\mathbf{92.30 \pm .16}$ & $85.10 \pm .27$ & $88.05 \pm .11$ & $88.78 \pm .33$ & $\color{darkblue} \mathbf{89.72 \pm .21}$ & $\mathbf{92.93 \pm .24}$ \\
& 200 & $92.03 \pm .10$ & $92.02 \pm .23$ & $\color{darkblue}  \mathbf{92.22 \pm .18}$ & $92.20 \pm .03$ & $\mathbf{93.66 \pm .33}$ & $88.18 \pm .16$ & $90.19 \pm .32$ & $90.67 \pm .18$ & $\color{darkblue} \mathbf{91.19 \pm .20}$ & $\mathbf{94.16 \pm .12}$ \\
\hline

    \end{tabular}
        }

    \end{sidewaystable}

    \begin{sidewaystable}
    \caption{CIFAR-10 results for ConvMixer-256/8 with patch size 1.   \textbf{Bold} denotes the highest per group,
        and {\color{darkblue} \textbf{blue bold}} denotes the second highest.    }
    \label{fig:apx-cm-256-8-1}
    \vspace{0.5em}
        \centering
    {\scriptsize
    \begin{tabular}{|c|c|b|b|b|b|b|a|a|a|a|a|}
        \hline
         &  & \multicolumn{5}{b|}{\textsc{Thawed}} & \multicolumn{5}{a|}{\textsc{Frozen}} \\
        \hline
        \makecell{Filt. \\ Size} & \makecell{\# \\ Eps} & Uniform & Cov. transfer & \makecell{Cov. transfer \\ ($\tfrac{1}{8}$ width)} & Direct transfer & Our init & Uniform & Cov. transfer & \makecell{Cov. transfer \\ ($\tfrac{1}{8}$ width)} & Direct transfer & Our init \\
        \hline
        \multirow{3}{*}{3} & 20 & $90.41 \pm .11$ & $90.60 \pm .31$ & $\mathbf{90.78 \pm .22}$ & $\color{darkblue}  \mathbf{90.66 \pm .18}$ & $89.84 \pm .27$ & $89.39 \pm .09$ & $\color{darkblue} \mathbf{90.37 \pm .06}$ & $90.14 \pm .03$ & $\mathbf{90.64 \pm .17}$ & $89.07 \pm .21$ \\
& 50 & $91.89 \pm .06$ & $92.00 \pm .23$ & $\mathbf{92.09 \pm .18}$ & $\color{darkblue}  \mathbf{92.05 \pm .20}$ & $91.49 \pm .10$ & $90.97 \pm .19$ & $\color{darkblue} \mathbf{91.87 \pm .09}$ & $91.65 \pm .10$ & $\mathbf{91.90 \pm .26}$ & $90.73 \pm .26$ \\
& 200 & $92.16 \pm .19$ & $92.38 \pm .12$ & $\mathbf{92.58 \pm .09}$ & $\color{darkblue}  \mathbf{92.54 \pm .18}$ & $91.85 \pm .15$ & $91.71 \pm .20$ & $\mathbf{92.12 \pm .08}$ & $\color{darkblue} \mathbf{92.05 \pm .36}$ & $91.98 \pm .16$ & $91.47 \pm .14$ \\
\hline
\multirow{3}{*}{7} & 20 & $91.70 \pm .16$ & $91.94 \pm .08$ & $\color{darkblue}  \mathbf{92.02 \pm .05}$ & $\mathbf{92.37 \pm .07}$ & $91.80 \pm .23$ & $89.84 \pm .04$ & $90.78 \pm .27$ & $\color{darkblue} \mathbf{91.30 \pm .11}$ & $\mathbf{91.92 \pm .03}$ & $91.08 \pm .15$ \\
& 50 & $93.30 \pm .15$ & $93.00 \pm .25$ & $93.31 \pm .10$ & $\mathbf{93.42 \pm .07}$ & $\color{darkblue}  \mathbf{93.33 \pm .29}$ & $91.51 \pm .20$ & $92.21 \pm .06$ & $\color{darkblue} \mathbf{92.68 \pm .07}$ & $\mathbf{93.01 \pm .02}$ & $92.31 \pm .32$ \\
& 200 & $93.67 \pm .13$ & $93.59 \pm .08$ & $93.68 \pm .12$ & $\color{darkblue}  \mathbf{93.81 \pm .23}$ & $\mathbf{93.83 \pm .18}$ & $92.48 \pm .16$ & $92.89 \pm .18$ & $92.98 \pm .05$ & $\mathbf{93.38 \pm .04}$ & $\color{darkblue} \mathbf{93.16 \pm .20}$ \\
\hline
\multirow{3}{*}{9} & 20 & $91.92 \pm .16$ & $91.74 \pm .25$ & $92.07 \pm .19$ & $\color{darkblue}  \mathbf{92.21 \pm .17}$ & $\mathbf{92.37 \pm .19}$ & $89.45 \pm .02$ & $90.41 \pm .24$ & $91.25 \pm .16$ & $\mathbf{91.85 \pm .03}$ & $\color{darkblue} \mathbf{91.35 \pm .05}$ \\
& 50 & $93.25 \pm .23$ & $92.92 \pm .05$ & $93.22 \pm .16$ & $\color{darkblue}  \mathbf{93.37 \pm .01}$ & $\mathbf{93.45 \pm .19}$ & $91.13 \pm .13$ & $91.93 \pm .17$ & $\color{darkblue} \mathbf{92.52 \pm .03}$ & $\mathbf{92.85 \pm .16}$ & $92.45 \pm .27$ \\
& 200 & $93.74 \pm .35$ & $93.60 \pm .16$ & $93.68 \pm .14$ & $\color{darkblue}  \mathbf{93.83 \pm .03}$ & $\mathbf{94.12 \pm .24}$ & $91.96 \pm .07$ & $92.67 \pm .04$ & $93.08 \pm .17$ & $\color{darkblue} \mathbf{93.38 \pm .01}$ & $\mathbf{93.56 \pm .14}$ \\
\hline
\multirow{3}{*}{15} & 20 & $90.65 \pm .23$ & $91.00 \pm .05$ & $\color{darkblue}  \mathbf{91.91 \pm .17}$ & $91.70 \pm .28$ & $\mathbf{92.29 \pm .21}$ & $86.64 \pm .14$ & $89.09 \pm .32$ & $90.70 \pm .07$ & $\color{darkblue} \mathbf{91.32 \pm .10}$ & $\mathbf{91.53 \pm .12}$ \\
& 50 & $92.76 \pm .06$ & $92.54 \pm .10$ & $\color{darkblue}  \mathbf{92.99 \pm .13}$ & $92.95 \pm .10$ & $\mathbf{93.42 \pm .07}$ & $89.20 \pm .17$ & $90.65 \pm .15$ & $91.72 \pm .23$ & $\color{darkblue} \mathbf{92.35 \pm .13}$ & $\mathbf{92.73 \pm .07}$ \\
& 200 & $93.55 \pm .16$ & $93.22 \pm .16$ & $93.57 \pm .03$ & $\color{darkblue}  \mathbf{93.59 \pm .02}$ & $\mathbf{94.13 \pm .13}$ & $90.20 \pm .22$ & $91.83 \pm .14$ & $92.79 \pm .17$ & $\color{darkblue} \mathbf{93.01 \pm .27}$ & $\mathbf{93.47 \pm .05}$ \\
\hline

    \end{tabular}
        }

    \end{sidewaystable}

\clearpage

%%%%%%%%%%%%%%%%%%%%%%%%%%%%%%%%%%%
% convnext

\begin{sidewaystable}
    \caption{CIFAR-10 results for ConvNeXt-atto with patch size 1. \textbf{Bold} denotes the highest per group,
        and {\color{darkblue} \textbf{blue bold}} denotes the second highest.  }
    \label{fig:apx-cm-256-1-1}
    \vspace{0.5em}
        \centering
    {\scriptsize
    \begin{tabular}{|c|c|b|b|b|b|b|a|a|a|a|a|}
        \hline
         &  & \multicolumn{5}{b|}{\textsc{Thawed}} & \multicolumn{5}{a|}{\textsc{Frozen}} \\
        \hline
        \makecell{Filt. \\ Size} & \makecell{\# \\ Eps} & Uniform & Cov. transfer & \makecell{Cov. transfer \\ ($\tfrac{1}{8}$ width)} & Direct transfer & Our init & Uniform & Cov. transfer & \makecell{Cov. transfer \\ ($\tfrac{1}{8}$ width)} & Direct transfer & Our init \\
        \hline
        \multirow{3}{*}{7} & 20 & $81.46 \pm .51$ & $85.00 \pm .21$ & $84.71 \pm .25$ & $\color{darkblue}  \mathbf{86.18 \pm .59}$ & $\mathbf{86.44 \pm .78}$ & $80.45 \pm .88$ & $83.91 \pm .13$ & $84.06 \pm .19$ & $\mathbf{86.12 \pm .57}$ & $\color{darkblue} \mathbf{84.71 \pm .08}$ \\
& 50 & $87.55 \pm .46$ & $88.38 \pm .10$ & $87.92 \pm .27$ & $\color{darkblue}  \mathbf{89.07 \pm .75}$ & $\mathbf{90.54 \pm .16}$ & $85.18 \pm .39$ & $87.81 \pm .10$ & $86.92 \pm .37$ & $\color{darkblue} \mathbf{89.32 \pm .21}$ & $\mathbf{90.13 \pm .14}$ \\
& 200 & $90.47 \pm .68$ & $90.77 \pm .46$ & $90.69 \pm .11$ & $\color{darkblue}  \mathbf{91.05 \pm .40}$ & $\mathbf{92.03 \pm .35}$ & $87.10 \pm .08$ & $89.70 \pm .21$ & $89.08 \pm .24$ & $\color{darkblue} \mathbf{90.83 \pm .21}$ & $\mathbf{92.27 \pm .22}$ \\
\hline

    \end{tabular}
        }

        \bigskip
        \bigskip
        \bigskip
        \bigskip
        \bigskip

        \caption{CIFAR-10 results for ConvNeXt-atto with patch size 2. \textbf{Bold} denotes the highest per group,
        and {\color{darkblue} \textbf{blue bold}} denotes the second highest.}
    \label{fig:apx-cm-256-1-2}
    \vspace{0.5em}
        \centering
    {\scriptsize
    \begin{tabular}{|c|c|b|b|b|b|b|a|a|a|a|a|}
        \hline
         &  & \multicolumn{5}{b|}{\textsc{Thawed}} & \multicolumn{5}{a|}{\textsc{Frozen}} \\
        \hline
        \makecell{Filt. \\ Size} & \makecell{\# \\ Eps} & Uniform & Cov. transfer & \makecell{Cov. transfer \\ ($\tfrac{1}{8}$ width)} & Direct transfer & Our init & Uniform & Cov. transfer & \makecell{Cov. transfer \\ ($\tfrac{1}{8}$ width)} & Direct transfer & Our init \\
        \hline
        \multirow{3}{*}{7} & 20 & $72.96 \pm .70$ & $\color{darkblue}  \mathbf{82.36 \pm .32}$ & $81.29 \pm .78$ & $\mathbf{83.42 \pm .47}$ & $81.13 \pm .54$ & $72.18 \pm .30$ & $\color{darkblue} \mathbf{80.82 \pm .17}$ & $79.03 \pm .07$ & $\mathbf{83.61 \pm .23}$ & $79.36 \pm .07$ \\
& 50 & $83.71 \pm .60$ & $86.21 \pm .21$ & $\mathbf{86.32 \pm .16}$ & $\color{darkblue}  \mathbf{86.28 \pm .20}$ & $85.97 \pm .24$ & $77.92 \pm .34$ & $85.16 \pm .45$ & $84.83 \pm .09$ & $\mathbf{85.93 \pm .20}$ & $\color{darkblue} \mathbf{85.45 \pm .20}$ \\
& 200 & $86.28 \pm .02$ & $87.40 \pm .27$ & $\color{darkblue}  \mathbf{88.14 \pm .18}$ & $87.02 \pm .10$ & $\mathbf{88.20 \pm .52}$ & $79.82 \pm .38$ & $86.25 \pm .10$ & $\color{darkblue} \mathbf{86.60 \pm .02}$ & $86.59 \pm .33$ & $\mathbf{87.69 \pm .28}$ \\
\hline

    \end{tabular}
        }

    \end{sidewaystable}

\end{document}